\pdfoutput=1
\documentclass[10pt,twocolumn,letterpaper]{article}

\usepackage[pagenumbers]{cvpr} 

\usepackage{color}
\usepackage{graphicx}
\usepackage{amsmath}
\usepackage{amssymb}
\usepackage{booktabs}
\usepackage{multirow}
\usepackage{float}
\usepackage{lipsum}

\usepackage[pagebackref,breaklinks,colorlinks]{hyperref}

\usepackage[capitalize]{cleveref}
\usepackage[accsupp]{axessibility}  
\def\cvprPaperID{11116} 
\def\confName{CVPR}
\def\confYear{2023}

\makeatother

\begin{document}
\title{CLIP${^2}$: Contrastive Language-Image-Point Pretraining from \\ Real-World Point Cloud Data}

\author{
Yihan Zeng\textsuperscript{\rm 1$^{*}$}, 
Chenhan Jiang\textsuperscript{\rm 2$^{*}$},
Jiageng Mao\textsuperscript{\rm 3},
Jianhua Han\textsuperscript{\rm 1},
Chaoqiang Ye\textsuperscript{\rm 1},\\
Qingqiu Huang\textsuperscript{\rm 1},
Dit-Yan Yeung\textsuperscript{\rm 2},
Zhen Yang\textsuperscript{\rm 1},
Xiaodan Liang\textsuperscript{\rm 4},
Hang Xu\textsuperscript{\rm 1$^\dagger$}\\
\vspace{-15mm}
}

\twocolumn[{%
\maketitle

\begin{figure}[H]
\hsize=\textwidth 
\centering
		\includegraphics[width=1\textwidth]{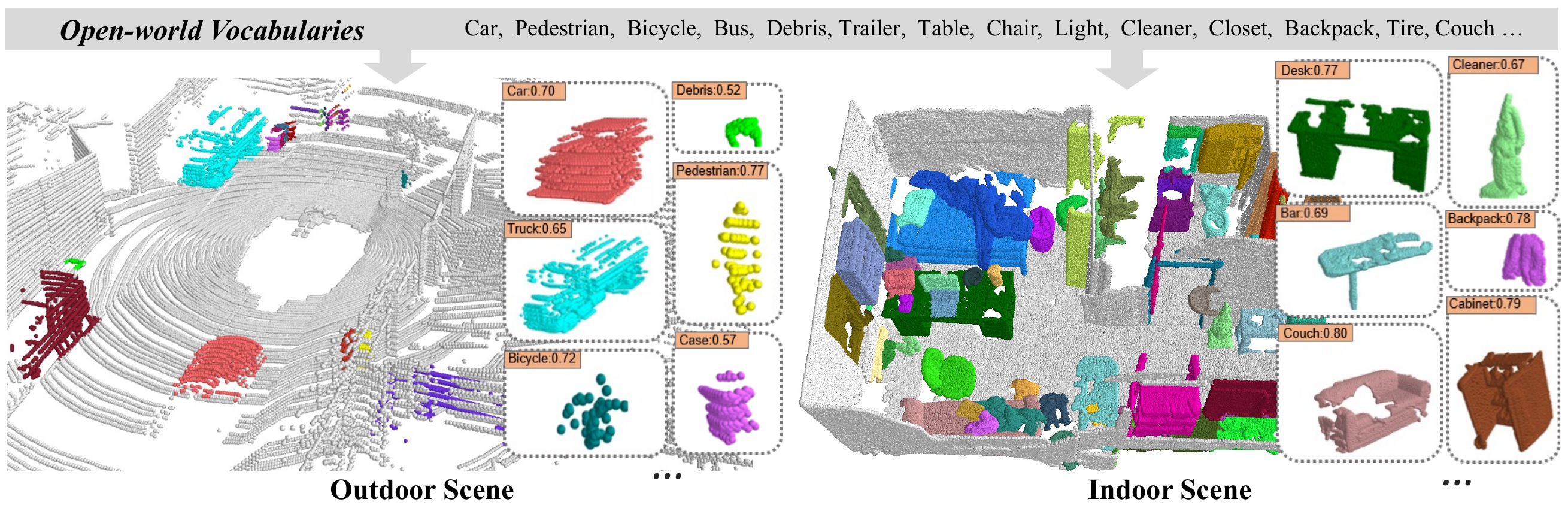}

		\vspace{-3mm}
	\caption{\textbf{Illustration of our open-world recognition results.} Benefiting from our CLIP$^2$, the 3D representation is aligned to the open-world language representation, which enables flexible zero-shot transfer. Best viewed in colors. 
	}
	\label{fig-intro}
\end{figure}

}]

\newcommand\blfootnote[1]{%
\begingroup
\renewcommand\thefootnote{}\footnote{#1}%
\addtocounter{footnote}{-1}%
\endgroup
}
\blfootnote{$^*$Equal contribution.  \hspace{4mm} $^1$Huawei Noah's Ark Lab \hspace{2mm} $^2$Hong Kong University of Science and Technology \hspace{2mm} $^3$The Chinese University of Hong Kong \hspace{2mm} $^4$Sun Yat-san University \hspace{2mm} $^\dagger$Corresponding Author: \url{xu.hang@huawei.com}}

\begin{abstract}
Contrastive Language-Image Pre-training, benefiting from large-scale unlabeled text-image pairs, has demonstrated great performance in open-world vision understanding tasks. 
However, due to the limited Text-3D data pairs, adapting the success of 2D Vision-Language Models (VLM) to the 3D space remains an open problem. 
Existing works that leverage VLM for 3D understanding generally resort to constructing intermediate 2D representations for the 3D data, but at the cost of losing 3D geometry information. To take a step toward open-world 3D vision understanding, we propose \textbf{C}ontrastive \textbf{L}anguage-\textbf{I}mage-\textbf{P}oint Cloud \textbf{P}retraining (CLIP$^2$) to directly learn the transferable 3D point cloud representation in realistic scenarios with a novel proxy alignment mechanism. 
Specifically, we exploit naturally-existed correspondences in 2D and 3D scenarios, and build well-aligned and instance-based text-image-point proxies from those complex scenarios.
On top of that, we propose a cross-modal contrastive objective to learn semantic and instance-level aligned point cloud representation.
Experimental results on both indoor and outdoor scenarios show that our learned 3D representation has great transfer ability in downstream tasks, including zero-shot and few-shot 3D recognition, which boosts the state-of-the-art methods by large margins.   
Furthermore, we provide analyses of the capability of different representations in real scenarios and present the optional ensemble scheme. 

\end{abstract}

\section{Introduction}

Powerful 3D point cloud representation plays a crucial role in various real-world applications, e.g., 3D object recognition and detection~\cite{yin2020center,shi2019pointrcnn, mao20223d, ding2019votenet,zhang2020h3dnet}. 
Compared to 2D images, 3D point cloud provides specific information like accurate geometry that is robust to illumination changes. 
However, current methods \cite{yin2020center,qi2017pointnet} that learn 3D representations generally rely on the predefined number of object categories and require plenty of labor-intensive annotations. 
Those learned 3D representations are insufficient for safety-critical scenarios like self-driving which includes
a long-tail class distribution far beyond the predefined taxonomy.
 Therefore, it is highly demanded to learn a transferable 3D representation equipped with zero-shot recognition ability in vocabulary scalable real-world scenes.  Figure ~\ref{fig-intro} shows an open-world recognition example by our CLIP$^2$ in outdoor and indoor scenes, where the 3D objects can be classified with the correlation alignment between 3D representations and open-world vocabularies.


The critical ingredient of open-world understanding is that the models learn sufficient knowledge to obtain general representations. To achieve this, recent Vision-Language Models (VLM) \cite{radford2021learning,jia2021scaling,yao2022detclip} leverage Internet-scale text-image pairs to conduct vision-language pretraining, which facilitates transferable 2D representation and demonstrates promising performance in 2D open-vocabulary tasks.
However,
3D vision-language pretraining remains unexplored due to the limitation
of existing 3D datasets in diversity and scale compared to 
the massive data sources in 2D counterparts \cite{li2021align, radford2021learning, jia2021scaling,yao2022detclip}. Though some recent works \cite{huang2022clip2point,zhang2022pointclip,ha2022semantic}
try to avoid this problem by transferring the pretrained 2D VLM into
the intermediate representation including projected image patches \cite{ha2022semantic,lu2022open}
or depth maps \cite{zhang2022pointclip,afham2022crosspoint}, 
those representations suffer from the loss of 3D geometric information
and limited viewpoints under realistic scenarios. Especially the camera images are only sometimes available due to the sensor failure in 3D scenes. We believe the 3D representation based on original point cloud data retains most information and is the optimal solution for 3D real world understanding, which requires a rethink of learning the transferable 3D representation under realistic scenarios.

To this end,
we propose a \textbf{C}ontrastive \textbf{L}anguage-\textbf{I}mage-\textbf{P}oint cloud \textbf{P}retraining framework, short for CLIP$^{2}$,  which directly aligns 3D space with broader raw text and advances the 3D representation learning into an open-world era.
Our learning process can be decomposed into two stages:
\textbf{Firstly,}  we introduce a \textit{Triplet Proxy Collection} to alleviate the limitation of accessible pretraining data by constructing language-image-point triplets from real-world scenes. Since the large-scale realistic 3D datasets for outdoor driving \cite{mao2021one,caesar2020nuscenes} and indoor scenarios \cite{dai2017scannet,song2015sun} are collected in open-world, it contains huge amounts of realistic objects that vary in semantics and diversity. Thus we consider them as potential pretraining data sources without extra human supervision. Specifically, we propose ``Proxy'' instances as the bridges between language descriptions, 2D images and 3D point clouds. Enabled by a well-aligned VLM, a scalable caption list and the geometry transformation between 2D and 3D, we automatically create more than 1 million triplets to facilitate pretraining. 
\textbf{Secondly,} we further propose a \textit{Cross-Modal Pretraining} scheme to jointly optimize the feature space alignments of three modalities, \textit{i.e.}point cloud, language and image. It contains both the contrastive learning objective of semantic-level text-3D correlation and instance-level image-3D correlation, which contributes to better transferability of learned 3D representation.

We study the transferable capability of CLIP${^2}$ by benchmarking the zero-shot recognition performance on four popular indoor and outdoor real-world datasets, and find a significant improvement over current methods, achieving Top1 accuracy 61.3\% on SunRGBD \cite{song2015sun}, 43.8\% on ScanNet \cite{dai2017scannet}), 28.8\% on nuScenes \cite{caesar2020nuscenes} and 56.0\% on ONCE \cite{mao2021one}. For a fair comparison with existing methods \cite{zhang2022pointclip, huang2022clip2point, afham2022crosspoint, wang2021unsupervised}, we conduct zero-shot and few-shot classification on single object dataset ScanObjectNN \cite{uy2019revisiting} and find consistent dominance, 16.1\% relative improvement on zero-shot classification over previous state-of-the-art method \cite{huang2022clip2point}. To validate the vocabulary-increasing ability of CLIP${^2}$, we report the quantity results and visualizations to show the improved discovery of the long-tail categories.
Moreover, we make ablations and analisis on different representations, and investigate ensembling alternatives to merge complementary knowledge of all available representations in realistic applications. Our contributions can be summarized as follows:

\begin{itemize}
    \item We propose a novel CLIP$^2$ framework that aligns 3D space with open-world language representation, facilitating zero-shot transfer in realistic scenarios.
    \item We present a Triplet Proxies Collection scheme in real-world scenes, which alleviates the shortage of text-3D data sources and facilitates the pretraining methods.
    \item CLIP$^2$ jointly optimizes the correlation alignment between point cloud, language and image by proposed cross-modal pretraining mechanism, which enhances the transferability of learned 3D representation.
    \item Our CLIP$^{2}$ achieves the state-of-the-art zero-shot transfer performance on 5 datasets (indoor/outdoor scenes and single-object) and shows quality results on vocabulary-increasing discovery in real world.
\end{itemize}

\section{Related Work}

\paragraph{Vision-Language Model. }
    Large vision language models (VLM)~\cite{li2021align, radford2021learning, jia2021scaling,yao2022detclip}  have demonstrated successful performance in downstream zero-shot tasks with the learned transferable 2D representations. CLIP \cite{radford2021learning} and ALIGN \cite{jia2021scaling} push the limit by collecting Internet-scale image-text pairs and then learning the correlation alignment between image and language feature space with contrastive pretraining objectives. Those models can be directly transferred to zero-shot 2D recognition and achieve impressive results. Recent DetClip~\cite{yao2022detclip} learns to align image patches to test phrases after pretraining under hybrid supervision from detection, grounding and image-text pair data, which extends the ability to localize open-vocabulary 2D proposals in images. In this paper, we  attempt to transfer the open-vocabulary ability of pre-trained VLM to the 3D domain, making language applicable to zero-shot point cloud recognition.
\begin{figure*}[t]
	\begin{center}
		\includegraphics[width=1.0\linewidth]{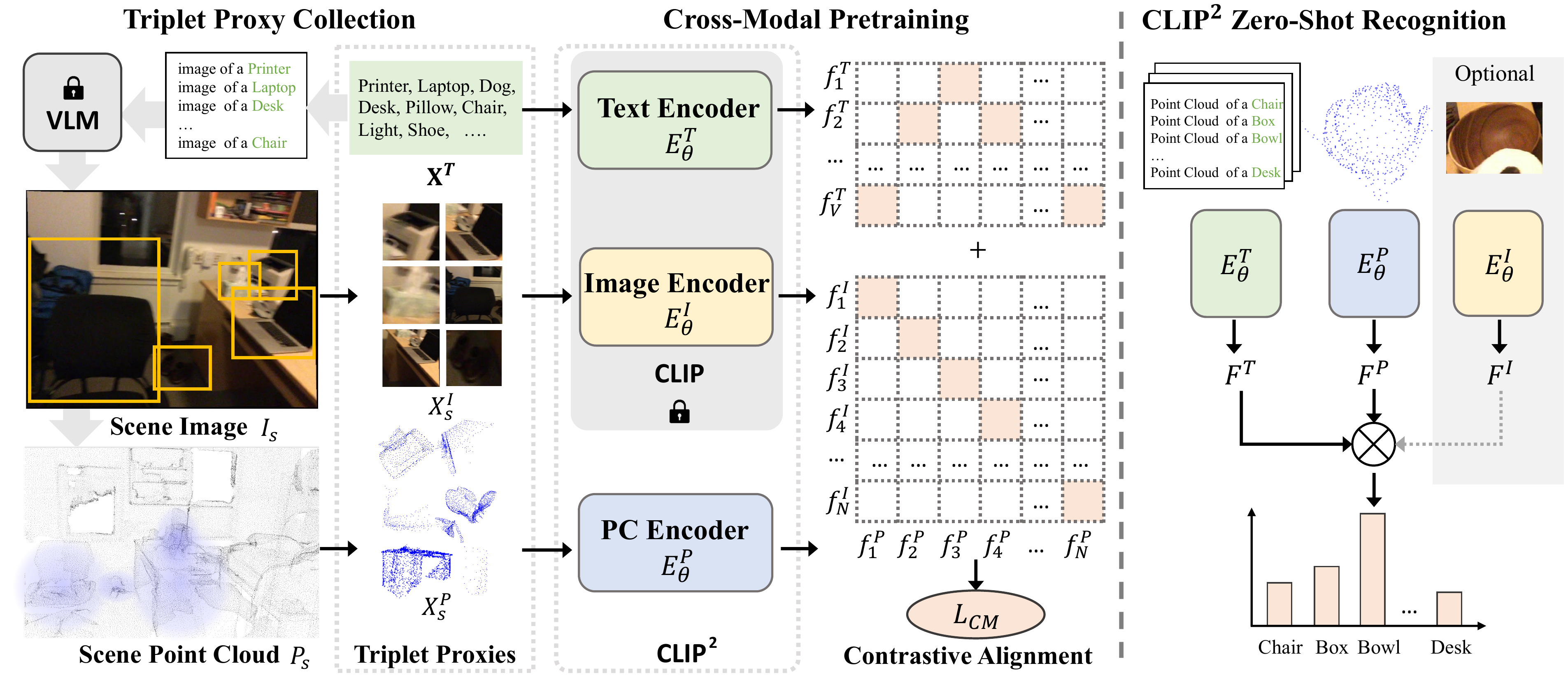}
	\end{center}
	\vspace{-3mm}
	\caption{\textbf{Overview of CLIP${^2}$ framework.} The main components contain two parts, the \textit{Triplet Proxy Collection} and the \textit{Cross-Modal Pretraining}. The defined Triplet Proxy set $\mathcal{D}_{\text{proxy}}$ consists of language captions ${\textbf{\text{X}}^T}$, corresponding image instances ${\textbf{\text{X}}^I}$ and raw 3D point cloud instances ${\textbf{\text{X}}^P}$, which come from the free data source under realistic scenarios without any labeling labor. On top of that, we pretrain a point cloud encoder ${E^P}$ with the cross-modal contrastive learning objective. Equipped with CLIP${^2}$, the learned 3D point cloud representation $F^P$ is well aligned to the language representation, which facilitates downstream zero-shot 3D transfer tasks in the real world. 
	}
	\label{fig-model}
\end{figure*}

\vspace{1mm}
\noindent\textbf{Zero-shot/Open-world Learning in 3D.}
Recognizing 3D objects with a large vocabulary is necessary for safety-critical autonomous driving and robotic tasks, yet remains under-explored. Cheraghian et al. \cite{cheraghian2019zero,cheraghian2019mitigating,cheraghian2022zero,cheraghian2020transductive} first attempt to associate PointNet \cite{qi2017pointnet} feature with category semantic information via a projection function, and separately proposed an unsupervised skewness loss \cite{cheraghian2019mitigating} to mitigate the hubness problem. The transductive case \cite{cheraghian2020transductive}
is discussed in which extends \cite{cheraghian2019mitigating} using a triplet loss. Notably, the above works conduct experiments on synthetic datasets and need to divide datasets into ``seen'' categories as training data and
``unseen'' categories as testing data. Thus they are not suitable for realistic scenarios due to the domain gap between synthetic and real-world data, as well as the limited vocabulary-increasing ability. Recently, inspired by the success of VLMs\cite{radford2021learning,jia2021scaling} in 2D tasks, some works~\cite{zhang2022pointclip, huang2022clip2point} propose to transfer the zero-shot recognition ability of pretrained CLIP~\cite{radford2021learning} into 3D area. PointCLIP \cite{zhang2022pointclip} directly projects point cloud into multi-view depth maps as image-like data input for pretrained CLIP to make classification predictions. While CLIP2Point~\cite{huang2022clip2point} trains an image-depth embedding on ShapeNet \cite{ye2020sarpnet} to better align the depth representation to the pretrained image space of CLIP. However, depth maps lost plenty of geometry information of the original point cloud data structure, resulting in poor performance especially in realistic scenarios. By contrast, we aim to learn transferable 3D representation based on the original point cloud data structure in realistic scenarios. 

\vspace{1mm}
\noindent\textbf{3D Representation Learning. }
Much progress has been made in learning a comprehensive 3D representation in an unsupervised manner. Most works \cite{afham2022crosspoint,sharma2020self,xie2020pointcontrast,liang2021exploring,pang2022masked, liu2022masked,zhang2022point,yu2022point} follow the paradigm that conducts pretraining on unlabeled datasets and then finetunes on the limited downstream annotations. Though the improved transferability of 3D representation, they can not be directly transferred to zero-shot tasks with open-world vocabularies. In this work, we conduct language-image-point cloud pretraining, which learns transferable 3D representation aligned to open-vocabulary language space to facilitate the zero-shot transfer.

\section{Method}
In this section, we introduce CLIP${^2}$ to learn a transferable 3D point cloud representation with arbitrary category recognition ability under realistic scenarios, illustrated in Figure \ref{fig-model}. We will first present the \textit{Triplet Proxy Collection} in Section~\ref{sec:tp}, which utilizes a pretrained VLM and geometric transformation to obtain language-image-point triplets from real-world scenes. Then we will elaborate \textit{Cross-Modal Contrastive Pretraining} mechanism in Section~\ref{sec:learn}, which jointly optimizes the alignment correlations between language, image and point cloud feature space. 

\subsection{Triplet Proxy Collection} \label{sec:tp}

Inspired by the significant performance of 2D VLMs on open-vocabulary tasks, we aim to develop 3D vision-language pretraining to facilitate category-increasing capacity for real-world scenarios. 
However, the core challenge is the shortage of pretraining data. 
Compared to the 2D vision-language pretraining framework CLIP~\cite{radford2021learning}, which takes more than 400M image-language pairs from the Internet, the largest 3D single-object dataset ShapeNet \cite{ye2020sarpnet} only contains 50K CAD models with 55 categories. In addition to the insufficiency of data scale,  pretraining on such synthetic data fails to transfer well in the real world due to the huge domain gap. 
Enlightened by the recent emergence of large-scale point cloud datasets collected in indoor~\cite{song2015sun, dai2017scannet} and outdoor scenarios~\cite{mao2021one,caesar2020nuscenes}, we observe that those naturally-collected datasets potentially contain vast amounts of open-world objects that vary in semantics and diversity. Considering the data collection itself is cheap except for laborious annotation, we novelly take leverage of those available datasets without human annotations as a practical yet effective pretraining data source.

Specifically, given the realistic scene data $\mathcal{S} = \{(P_s, I_s)_{s=1}^{|\mathcal{S}|}\}$, where  $P_s\in \mathbb{R}^{N_P \times 3}$ and  $I_s\in \mathbb{R}^{N_I\times H\times W \times 3}$ are corresponding 3D point clouds and images of scene $s$, we propose a novel concept, \textit{Proxy}, as the bridge between language, image and 3D point cloud. As illustrated in Figure~\ref{fig-model}, equipped by those proxy instances, we can automatically collect a massive number of language-image-point cloud pairs $\mathcal{D}_{\text{proxy}}$ in the format of proxies under open-world scenes. We detail the process as follows.

\begin{figure}[t]
	\begin{center}
		\includegraphics[width=1.0\linewidth]{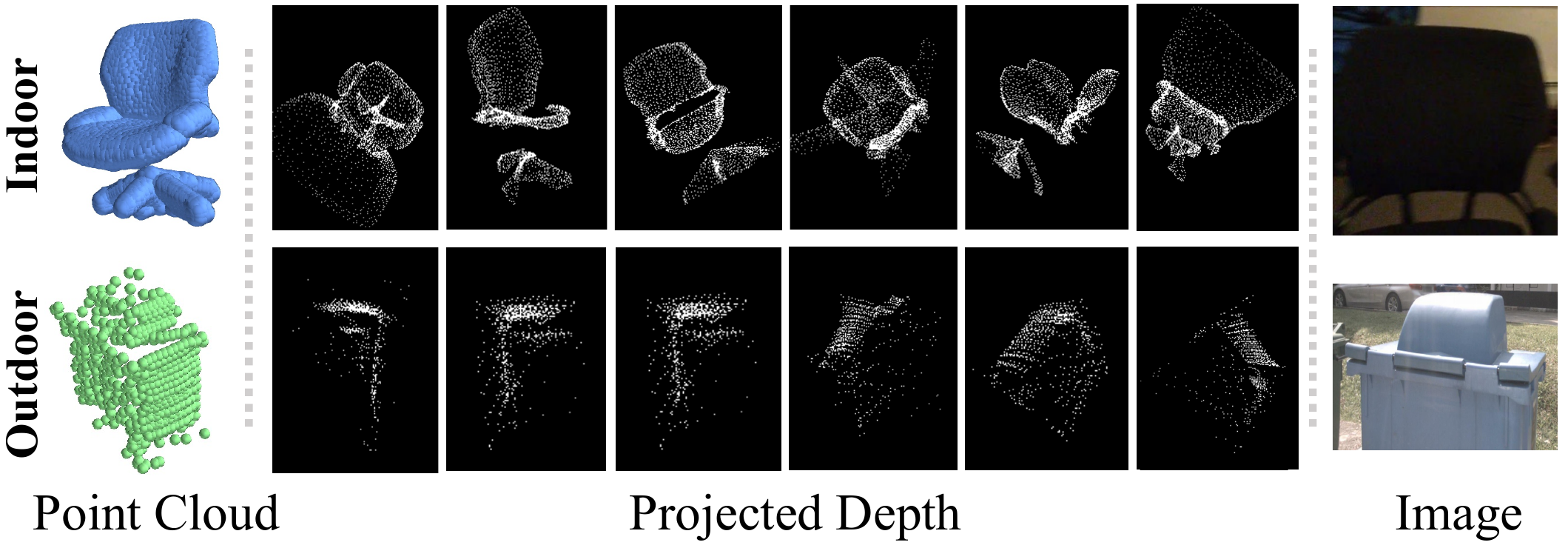}
	\end{center}
				\vspace{-4mm}
	\caption{\textbf{Illustration of three representation modals} of two 3D objects examples under indoor and outdoor scenarios. 
	}
	\label{fig-depth}
\end{figure}

\vspace{1mm}
\noindent\textbf{Language Proxy. }We set the language proxies $\textbf{\text{X}}^T\in \mathbb{R}^{V}$ as a raw text list from the 2D open-world dataset~\cite{gupta2019lvis}, where $V=1206$ denotes the vocabulary size of language proxies. 

\vspace{1mm}
\noindent\textbf{Image Proxy. }Next, we obtain the image proxies $\textbf{\text{X}}^I$ by an open vocabulary detector DetCLIP \cite{yao2022detclip}, denoted as $M$, which is trained with open-world data and performs open-set detection.
Concretely, given language proxies $\textbf{\text{X}}^T$ and input scene image $I_s$, we extract corresponding image proposals as image proxies $X^I_s$  with $M$ by the similarity between input language embeddings and proposal features as 
\begin{equation}
\{X^I_s\}_{s\in |\mathcal{S}|} = \text{M}(\{I_s\}_{s\in |\mathcal{S}|}, \textbf{\text{X}}^T). 
\end{equation} 

\vspace{1mm}
\noindent\textbf{3D Proxy. } We exploit the naturally-existed geometry relations between 2D and 3D scenes to obtain 3D proxies $\textbf{\text{X}}^P$, which consists of point cloud instances corresponding to image proposals in $\textbf{\text{X}}^I$. We simplify the geometry transformation as $\text{G}(\cdot)$ and formulate the relations as:
\begin{equation}
X^P_i =  \text{G}(X^I_i).
\end{equation}
Detailedly, for \textit{indoor scenes} equipped with RGB-D sensors,  
we first remove the background pixels by unsupervised segmentation algorithm~\cite{rother2004grabcut} for each image proxy ${X^I_{s, i}}$, $i\in|X^I_{s}|$. 
Since depth information is known, 
we then transform the segmented pixels from $uvd$ coordinate $X^{I,uvd}_{s,i}\in\mathbb{R}^{n, 3}$ to $xyz$ coordinate $X^{P, xyz}_{s,i}\in\mathbb{R}^{n, 3}$ as a 3D point cloud proxy with the given camera parameters.
For \textit{outdoor scenes} captured by LiDAR sensors, we first create a 3D frustum for each image proxy by extruding the 2D image proposal into 3D space following \cite{paigwar2021frustum,qi2018frustum}. Then we conduct DBSCAN algorithm~\cite{schubert2017dbscan} within frustum and select the point cloud cluster as the point proxy $X^{P, xyz}_{s, i}$.

Eventually, we construct Triplet Proxy $\mathcal{D}_{\text{proxy}}=\{\textbf{\text{X}}^T,X_{s}^{I},X_{s}^{P}\}_{s=1}^{|\mathcal{S}|}$ by combining corresponding language proxies ${\textbf{\text{X}}^T}$, image proxies $\textbf{\text{X}}^I$ and 3D proxies $\textbf{\text{X}}^P$, where $\textbf{\text{X}}^I$=$\{{X_{s}^{I}}\}_{s=1}^{|\mathcal{S}|}$  and $\textbf{\text{X}}^P$=$\{{X_{s}^{P}}\}_{s=1}^{|\mathcal{S}|}$. 220K and 1.4M proxy triplets are formed for indoor and outdoor scenes, respectively. More details can be found in the appendix. 

\begin{figure}[t]
	\begin{center}
		\includegraphics[width=1.0\linewidth]{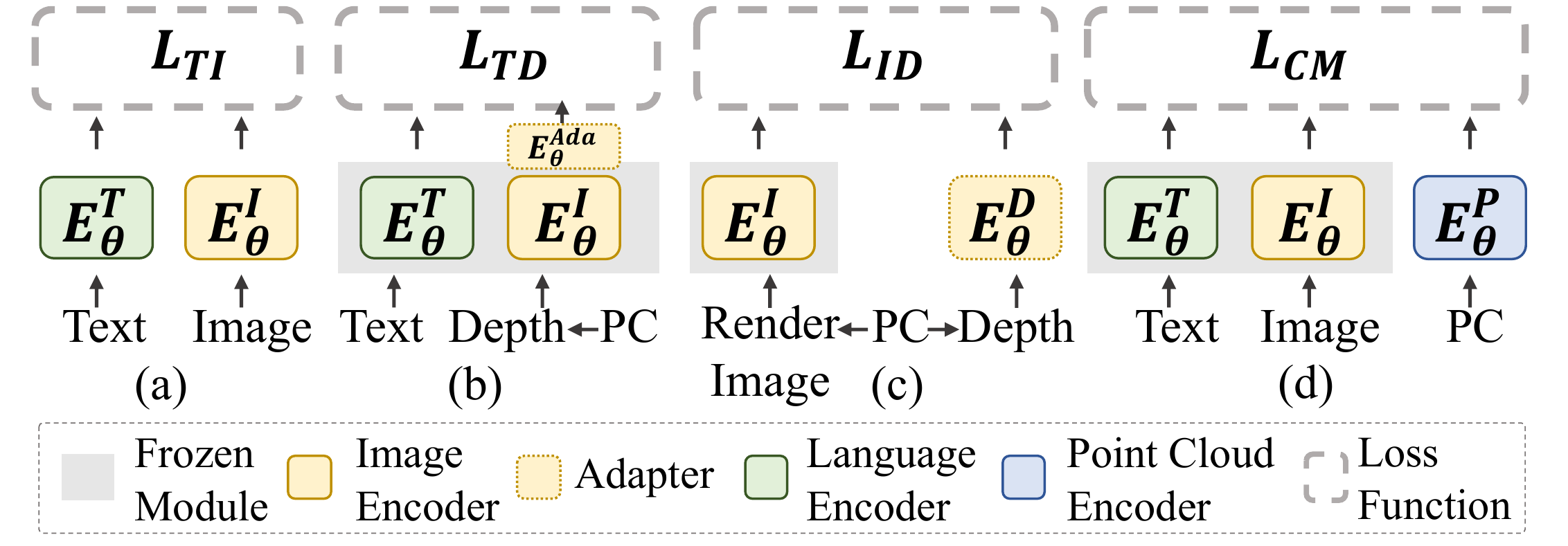}
	\end{center}
		\vspace{-4mm}
	\caption{\textbf{Comparison of different pretraining strategies.} \textbf{(a)} CLIP aligns image and language embedding space~\cite{radford2021learning} as $L_{TI}$ based on large-scale text-image pairs. \textbf{(b)} PointClip~\cite{radford2021learning} aligns projected depth map to CLIP language space as $L_{TD}$. \textbf{(c)} Clip2Point aligns depth map to CLIP image space as $L_{ID}$. \textbf{(d)} our CLIP$^2$ aligns original 3D point cloud to both CLIP language space and image space via cross-modal objective $L_{CM}$.
	}
	\label{fig-model2}
\end{figure}

\subsection{Cross-Modal Contrastive Pretraining}\label{sec:learn}

With the triplet proxies $\mathcal{D}_{\text{proxy}}$, a straightforward pretraining objective is forcing the alignment between the embedding spaces of point cloud ${X^P_i}$ and language ${X^T_i}$ from scratch. 
However, it might not promise good transferability of learned representation, since the number of language-image-point pretraining data triplets remains two orders of magnitude smaller than the language-image pairs adopted by CLIP \cite{radford2021learning} and the vocabulary size is much more limited. 
Therefore, we design to learn the correlation alignment based on the pretrained embedding space of CLIP.

The comparison of current pretraining strategies~\cite{zhang2022pointclip,huang2022clip2point} is illustrated in Figure \ref{fig-model2}, which is a series of 3D variants of CLIP. Notably, both existing methods exploit projected depth map as the intermediate representation of point cloud, which are respectively learned to align to language space~\cite{zhang2022pointclip} and image space \cite{huang2022clip2point}. Intuitively, as illustrated in Figure~\ref{fig-depth}, depth representation lost plenty of geometry information compared to the original point cloud, especially in outdoor scenarios. Moreover, images are sometimes unavailable for 3D objects. Thus we conduct pretraining on original 3D point cloud data as an optimal representation.

Toward learning more transferable representation, we introduce a cross-modal contrastive learning objective to jointly optimize the correlation alignment across language, image and point cloud, including \textit{Semantic-Level Language-3D Alignment} and \textit{Instance-Level Image-3D Alignment}. Specifically, the overall architecture of CLIP${^2}$, shown in Figure \ref{fig-model}, contains language encoder $E_{\theta}^T$, point cloud encoder $E_{\theta}^P$ and visual encoder $E_{\theta}^I$, which respectively embed the triplet proxies into text feature $f^T\in\mathcal{R}^{1\times C_T}$, point cloud feature $f^P\in\mathcal{R}^{1\times C_P}$ and image feature $f^I\in\mathcal{R}^{1\times C_I}$, where $C$ is the embedding dimension.

\vspace{1mm}
\noindent\textbf{Semantic-Level Language-3D Alignment. } In order to inherit the open-world recognization ability from pretrained CLIP \cite{radford2021learning}, we align the point cloud feature $f^P$ with text embedding $f^T$ from well-trained CLIP with Language-Point Proxy $\{X^{T}_i,X^{P}_i\}$ input. We replace \textit{classname} in the prompts, like ``point cloud of a \{\ \textit{classname} \}\ .” with raw text in proxy $X^{T}_i$ as language sentences. 
The core idea is to drive the feature centroids of 3D instances and the corresponding text prompt closer. We compute the loss function of between of language proxy and point cloud proxy as:
\vspace{1mm}
\begin{equation}\footnotesize
      l(i,T,P) = -\log\frac{\exp(f^T_i\cdot f^P_{i}/\tau)}{\exp(f_i^T\cdot f^P_i/\tau) + \sum\limits_{j\in N, X^T_j\neq X^T_i} \exp(f_i^T\cdot f_j^P/\tau)},
\end{equation}
where $N$ is the mini-batch size, $\tau$ is the temperature co-efficient. Within a training mini-batch, the language-3D alignment objective $L(T,P)$ can be described as:
\begin{equation}\small
       L(T,P) = \frac{1}{N}\sum_{i\in N} l(i, T, P).
\end{equation}

\vspace{1mm}
\noindent\textbf{Instance-Level Image-3D Alignment. } 
In addition to the alignment between semantic language and 3D proxy instances, we further introduce the contrastive alignment between instance-wise image proxy and 3D proxy instances. Note that the instance-aware visual concept has been well-studied in the embedding space of pretrained CLIP. We believe instance-sensitive learning can contribute to further correlation learning and benefits to the transferability of learned 3D representation.
The contrastive aligned objective $L(I,P)$ across point cloud and image is formulated as:
\begin{equation}\small
      l(i,I,P) = -\log\frac{\exp(f^I_i\cdot f^P_{i}/\tau)}{\exp(f_i^I\cdot f^P_i/\tau) + \sum\limits_{j\in N, j\neq i} \exp(f_i^I\cdot f_j^P/\tau)} ,
\end{equation}
\begin{equation}
      L(I,P) = \frac{1}{N}\sum_{i\in N} l(i, I, P) . 
\end{equation}

Finally, we obtain the resultant cross-modal contrastive learning objective $L_{CM}(T,I,P)$ as the combination of $L(T,P)$ and $L(I,P)$, where both alignments of semantic-level text-3D correlation and instance-level image-3D correlation are injected:

\begin{equation}\small
       L_{CM}(T,I,P) =  \lambda_1 L(T,P) + \lambda_1 L(I,P),
\end{equation}
where the hyper-parameters $\lambda_1$ and $\lambda_2$ are both set to 0.5.

\vspace{2mm}
\section{Experiment}

In this section, we evaluate CLIP${^2}$ on realistic indoor and outdoor scenarios. We report the zero-shot transfer results on various datasets~\cite{song2015sun, dai2017scannet, caesar2020nuscenes, sun2020scalability, uy2019revisiting} and make further analysis on the designs of pretraining strategy.

\begin{table*}[t]
		\begin{center}
	\resizebox{1\textwidth}{!}{
	\begin{centering}
	\begin{tabular}{c|c|ccccccccccc}
\toprule
Method & Avg. & Bed &  Bookshelf & Chair & Desk & Sofa & Table & toilet & Bathhub & Dresser & Night Stand\tabularnewline
\midrule
PointClip~\cite{zhang2022pointclip}  & 11.5 & 0.0  & 94.0  & 0.0 & 0.0 & 0.0 & 14.7 & 0.0  & 0.0 &  6.1 & 0.0  \tabularnewline
Clip2Point~\cite{huang2022clip2point} & 18.6 & 10.9 & 20.6 & 64.3 & 34.4 & 13.8 & 14.1 & 26.2 & 0.0 & 1.4 & 0.0   \tabularnewline
PointClip~\cite{zhang2022pointclip} w/ TP.  & 38.0 & 45.3 & 100.0  & 62.5  & 48.5 & 44.4  & 4.8 & 55.2  & 16.3  & 3.3  & 0  \tabularnewline
Clip2Point~\cite{huang2022clip2point} w/ TP. & 56.9 & 78.0  & 87.6  & 36.2 & 36.6 & 64.7 & 37.4 & 82.1 & 77.5 & 67.6  & 1.2   \tabularnewline
\midrule
CLIP${^2}$ & \textcolor{blue}{61.3} & 84.0  & 75.5  & 70.7 & 47.3 & 75.5 & 33.8 & 86.2 & 65.3 & 71.8 & 2.4 \tabularnewline
CLIP${^2}$ w/ En. & \textcolor{blue}{69.6} & 87.3  & 94.3  & 70.7 & 54.1 & 79.3 & 47.0 & 91.7 & 85.7 & 82.2 & 3.6 
\tabularnewline
\bottomrule
\end{tabular}
\end{centering}
		}
	\end{center}
	
	\vspace{-5mm}
		\caption{\textbf{Zero-shot recognition results in SUN RGB-D.} \textbf{Avg.: }the mean average Top1 accuracy across all categories. \textbf{TP.: }our Triplet Proxy set.  \textbf{En.: }our optional ensembling scheme for inference. 
	}\label{tab-main-sun}
\end{table*}

\begin{table*}[t]
	\begin{center}
	\resizebox{\textwidth}{!}{
	\begin{centering}
	\begin{tabular}{c|c|ccccccccccccccccc}
\toprule

Method & Avg. & Cab & Bed & Chair & Sofa & Tabl & Door & Wind & Bksf & Pic & Cntr & Desk& Curt & Fridg & Bath & Showr & Toil & Sink  \tabularnewline
\midrule
PointClip~\cite{zhang2022pointclip} & 6.3 & 0.0  & 0.0 & 0.0 &0.0  & 0.7 &0.0  & 0.0 & 91.8 & 0.0 &0.0 &0.0  &15.0  &0.0  &0.0 &0.0 &0.0 &0.0  \tabularnewline
Clip2Point~\cite{huang2022clip2point} & 24.9  & 0.0 & 20.8 & 85.1 & 43.3 & 26.5 & 69.9 & 0.0& 20.9 & 1.7 & 31.7 & 27.0 & 0.0 & 1.6 & 46.5&0.0&22.4&25.6    \tabularnewline
PointClip~\cite{zhang2022pointclip} w/  TP. &26.1&55.7&0.0&72.8&5.0&5.1&1.7&0.0&77.2&0.0&0.0&51.7&0.3&0.0&0.0&40.3&85.3&49.2\tabularnewline
Clip2Point~\cite{huang2022clip2point} w/  TP. & 35.2 &3.0 &11.8&45.1&27.6&10.5&61.5&2.6&71.9&0.3&33.6&29.9&4.7&11.5&72.2&92.4&86.1&34.0\tabularnewline
\midrule
CLIP${^2}$ & \textcolor{blue}{38.5} & 67.2& 32.6& 69.3 & 42.3 & 18.3 & 19.1 & 4.0 & 62.6 & 1.4 & 12.7 & 52.8 & 40.1 & 9.1 & 59.7 & 41.0 & 71.0 & 45.5  \tabularnewline
\bottomrule
\end{tabular}
\end{centering}
		}
	\end{center}

	\vspace{-5mm}
		\caption{\textbf{Zero-shot recognition in ScanNet.} \textbf{Avg.: }the mean average Top1 accuracy across all categories. \textbf{TP.: }our Triplet Proxy set. 
	}\label{tab-main-scan}
\end{table*}

\subsection{Zero-shot Transfer}
\paragraph{Setting. }

After pretraining, natural language is applied to reference learned 3D representation to enable following zero-shot transfer tasks. \textbf{(\romannumeral1) Zero-Shot Recognition: } we evaluate zero-shot recognition performance for realistic objects, where $K$ category names are transferred to text prompt \textquotedblleft point cloud of $\{$CLASS$\}$ \textquotedblright{} to encode the text features $F_K\in\mathbb{R}^{K\times C}$. Then the classification logits are calculated with the 3D feature $f^P$ and text features as:
\begin{equation}
    \text{logits}_i = \text{softmax}(f^P_i(F_K)^T).
\end{equation}
We present the results under both indoor and outdoor scenarios in Table~\ref{tab-main-sun}, Table~\ref{tab-main-scan} and Table~\ref{tab-main-out}, as well as the object-level benchmark in Table~\ref{tab-scanobj}. \textbf{(\romannumeral2) Open-vocabulary recognition: } we enlarge the category vocabularies of ScanNet to 249 and 384 to study the open-vocabulary recognition ability in Table~\ref{tab-ov}.  \textbf{(\romannumeral3) Open-vocabulary localization: } we study the open-vocabulary localization ability by localizing open-world 3D objects with our proxy generation process and then classifying them with our learned 3D representation, of which the visualization is illustrated in Figure~\ref{fig-vis} and evaluation results are reported in Table~\ref{tab-det}. Notably, we investigate representation ensembling alternatives to enable knowledge merging of all available representations for realistic applications, illustrated in Table~\ref{tab-ablation-ensemble}.


\subsubsection{Indoor Scenarios}
\paragraph{Datasets and details.}
We adopt the widely used indoor 3D dataset SUN RGB-D~\cite{song2015sun} as the realistic indoor scenario that provides pretraining data source, a single-view RGB-D dataset consisting of $\sim$10K scenes. To validate the  transferability of learned 3D representation, we also evaluate another popular indoor 3D dataset ScanNet~\cite{dai2017scannet}, which contains $\sim$1.5K scenes of 3D reconstructed meshes. 
We remove objects in ScanNet with less than 5 points, leaving 384 noisy categories.
For open-vocabulary recognition, we evaluate performance on the ScanNet 384-class set and a 249-class merged set.
In addition to the scene-wise indoor dataset, we conduct evaluations on ScanObjectNN~\cite{uy2019revisiting}, which collects $\sim$3K individual realistic objects with 15 categories and is applied in the previous zero-shot evaluation~\cite{zhang2022pointclip, huang2022clip2point}. During the proxy collection process, we empirically set $\epsilon=0.3$ in~\cite{yao2022detclip} as a tradeoff between filtering FPs and preserving TPs to generate image proxies. Considering the occurrence frequencies of different indoor categories vary a lot, we adopt the class balance strategy~\cite{mmdetection} to mitigate the class imbalance. During pretraining process, we adopt~\cite{qi2017pointnet++} as point cloud encoder and  set the overall training epoch number to 100.

\vspace{-1mm}
\paragraph{Quantity results. } 
For zero-shot recognition task, we take two recent works as our baselines, \textit{i.e.}  PointClip~\cite{zhang2022pointclip} and Clip2Point~\cite{huang2022clip2point}, which study the zero-shot classification task on 3D object-level benchmarks~\cite{vishwanath2009modelnet, uy2019revisiting} by leveraging pretrained CLIP with projected depth maps. Focusing on the real-world scenarios, we conduct comparison not only on the realistic object-level~\cite{uy2019revisiting} as illustrated in Table~\ref{tab-scanobj} but also on the scene-level datasets shown in Table~\ref{tab-main-sun} and Table~\ref{tab-main-scan}, where the evaluation follows the common classes split in~\cite{ding2019votenet, zhang2020h3dnet} and reports the instance Top1 accuracy of each class. As shown in tables, our CLIP$^2$ can outperform baselines on all benchmarks by large margins. Besides, we apply our triplet proxy generation mechanism (TP.) to baseline methods, and achieve considerable improvements on SUN RGB-D and ScanNet by 26.5\%  and 19.8\% for PointClip, 38.3\% and 10.3\% for Clip2Point. On the one hand, the contrasts demonstrate the effectiveness of  our triplet proxies for open-world understanding. On the other hand, our learned 3D representation is superior in 3D object recognition by retaining more 3D-specific information than depth representation. Besides, we present the optional ensembling scheme (En.) when camera images are available, which can take advantage of multi-modal knowledge and further boost the performance by 8.3\%.  To further validate the open-vocabulary recognition ability, we conduct evaluation on a larger category set of ScanNet in Table \ref{tab-ov} and report the instance Top5 accuracy, which illustrates the superiority of our CLIP$^2$ when vocabulary increases. Beyond that, CLIP$^2$ is also equipped with zero-shot 3D localization ability by proxy generation. On indoor scenario SUN RGB-D, we compare with a SOTA indoor 3D detector 3DETR \cite{misra2021end} and a recent work OV3D~\cite{lu2022open} that studies open-vocabulary detection, where evaluation is conducted on the same ''unseen'' split in~\cite{lu2022open}. Since CLIP$^2$ do not fit the tight bounding boxes of point cloud instances, we estimate the maximum bounding box of proxies and GT instances to conduct evaluation following the same metrics mAP$_{25}$ and AR$_{25}$ in~\cite{lu2022open},  as shown in Table~\ref{tab-det}. Notably, compared to baseline works that train on ``seen'' 3D annotations and test on ``unseen'' categories, we have no access to any 3D annotations yet achieve comparable localization ability, which yields 5.3\% AR$_{25}$ improvement over OV3D~\cite{lu2022open}.
We further evaluate segmentation results  in Table~\ref{tab-det}.

\begin{table}[]
\begin{minipage}{1.0\textwidth}
     \begin{minipage}[t]{0.18\textwidth}
      	\resizebox{\textwidth}{!}{
    	    \begin{centering}
                    \begin{tabular}{c|cc}
                        \toprule
                        \multirow{2}*{Method} & \multicolumn{2}{c}{Top5 Acc.} \\
                            		    ~ & 384 cls. &  249 cls. \\
                        \midrule
                        \cite{zhang2022pointclip} & 0.3 & 0.4 \tabularnewline
                        \cite{huang2022clip2point} & 6.4 &  7.0  \tabularnewline
                        \midrule
                        CLIP${^2}$ & 22.0 & 31.7  \tabularnewline
                        \bottomrule
                    \end{tabular}
            \end{centering}
    		}
    		\vspace{-2mm}
          \caption{Recognition results of vocabulary expansion in ScanNet. 
          }\label{tab-ov}
      \end{minipage}
    \begin{minipage}[t]{0.28\textwidth}
         	\resizebox{\textwidth}{!}{
        	    \begin{centering}
                    \begin{tabular}{c|ccc|cc}
                        \toprule
                        \multirow{2}*{Method}  & \multicolumn{3}{c|}{IN}& \multicolumn{2}{c}{OUT}\\
                        ~ & mAP$_{25}$ & AR$_{25}$ & mIoU & P. &R. \\
                        \midrule
                        \cite{misra2021end} &1.3&-&-&-& -\tabularnewline
                        \cite{lu2022open}&13.0&37.7 &-& - &-   \tabularnewline
                        \midrule
                        CLIP${^2}$  & 12.7 & 43.0& 52.8 & 6.7 &87.4 \tabularnewline
                        \bottomrule
                    \end{tabular}
                \end{centering}
        		}
        		    		\vspace{-2mm}
          \caption{\textbf{Zero-shot localization results.} \textbf{IN: }indoor scenario SUN RGB-D. \textbf{OUT: }outdoor scenario nuScenes.}\label{tab-det}
    \end{minipage}

\end{minipage}
\end{table}

\begin{table*}[t]
	\begin{center}
	\resizebox{\textwidth}{!}{
		\begin{tabular}{c|c|cccccccccc|ccccc} 
\toprule
\multirow{2}*{Method} &\multirow{2}*{Avg.} & \multicolumn{10}{c|}{nuScenes} & \multicolumn{5}{c}{ONCE}\\
~ & ~  & Car & Truck & Bus & Ped. & Bicycle & Trailer & C.V. & Motor. & Barrier & T.C.& Car & Cyc. &Ped. &Truck &Bus\tabularnewline
\midrule
PointClip~\cite{zhang2022pointclip} & 11.7 &0.0   &0.0  & 0.0 & 29.1 &41.8  & 3.4& 0.0 & 0.1 & 42.5 &0.0& 0.0 &13.8 &79.2 &0.2&7.61 \tabularnewline
Clip2Point~\cite{huang2022clip2point} &  12.4 & 0.4 & 0.3 & 0.1 & 13.5 & 31.0 & 1.8 &9.3  & 1.5 & 66.2 &0.3 & 17.3& 11.8& 95.4& 35.7& 4.0 
\tabularnewline
PointClip~\cite{zhang2022pointclip} w/  TP. & 28.3  & 18.8  & 0 & 5.5 & 74.0  & 17.9 & 57.0 & 1.9 & 4.5 & 2.1  & 29.7 & 51.8 & 9.2 & 99.8 & 5.0 & 46.8 \tabularnewline
Clip2Point~\cite{huang2022clip2point} w/  TP. &  33.0 & 26.7 & 16.8 & 51.2 & 45.2 & 15.8 & 13.9 & 20.0 & 5.7  & 10.5 & 34.2 & 39.4 & 27.8 & 95.5 & 40.6 & 51.7  \tabularnewline
\midrule
CLIP${^2}$ & \textcolor{blue}{37.8} & 41.9 & 41.3 & 22.5 & 40.3 & 21.1 & 20.6 & 24.8 & 22.4 & 17.3 & 35.3 & 52.7 & 27.3 & 77.7 & 78.5 & 44.0  \tabularnewline
\bottomrule
		\end{tabular}
		}
	\end{center}
				\vspace{-5mm}
	\caption{\textbf{Zero-shot recognition results in outdoor scenario}: nuScenes (Left) and ONCE (Right). \textbf{TP.: }our Triplet Proxy set. \textbf{Avg.: }the mean average Top1 accuracy across all categories of two benchmarks. 
	}\label{tab-main-out}
 \vspace{0.5mm}
\end{table*}

\begin{figure*}[t]
	\begin{center}
		\includegraphics[width=1.0\linewidth]{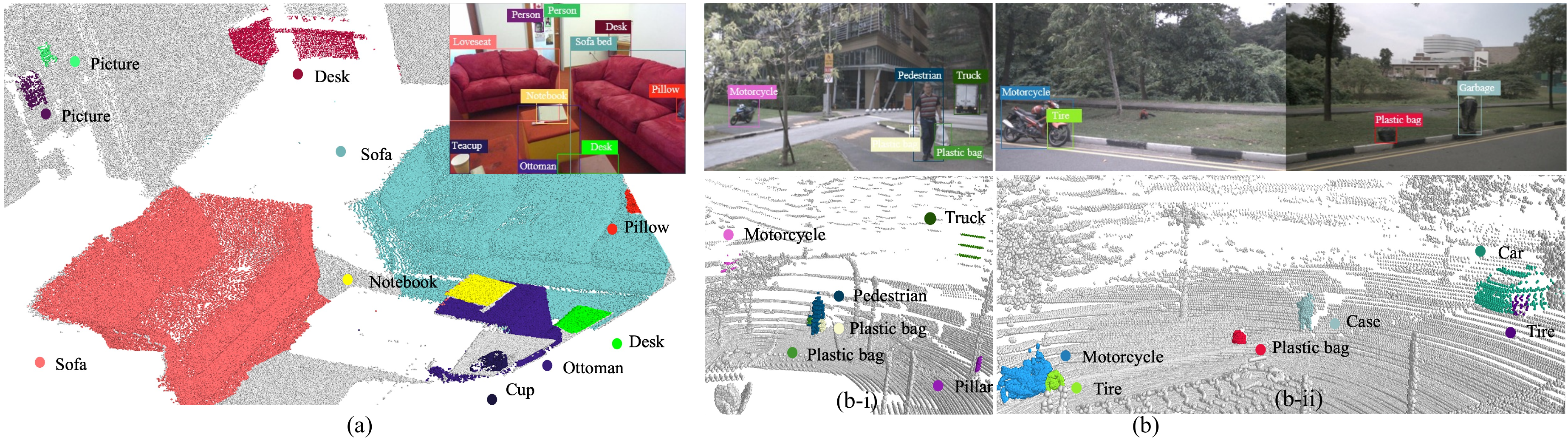}
	\end{center}
	\vspace{-6mm}
	\caption{\textbf{Visualizations of the zero-shot localization and recognition results} by  CLIP$^2$ under open-world \textbf{(a)} indoor realistic scene~\cite{song2015sun} and \textbf{(b)} outdoor scenes~\cite{caesar2020nuscenes}.
     Notably, the whole pipeline of CLIP$^2$ not only has no access to human annotations, but also enables the open-world vocabularies beyond groundtruth annotations, such as 'Picture' in \textbf{(a)} and ’Plastic bag', 'Tire' in \textbf{(b)}. Best viewed in colors. 
	} 
	\label{fig-vis}
\end{figure*}

\vspace{-5mm}
\paragraph{Quality results. } The visualization results of CLIP$^2$ under a indoor scene of SUN RGB-D~\cite{song2015sun} is shown in Figure~\ref{fig-vis}\textcolor{red}{(a)}. Our triplet proxy generation process can localize open-world 3D objects in a point cloud scene. Moreover, the 3D representation learned from our cross-modal pretraining can provide more accurate classification results for 3D instances by exploiting original point cloud, which corrects the mistaken ''People'' prediction in image to ''Picture'' by considering the geometry information.

\subsubsection{Outdoor Scenarios}
\paragraph{Datasets and details. }
We exploit a prevalent large-scale 3D dataset nuScenes~\cite{caesar2020nuscenes} as the outdoor data source and extra validate the performance on the ONCE dataset~\cite{mao2021one}. The nuScenes dataset consists of $\sim$28K frames with 10 categories, while ONCE contains 6 annotated sequences with 5 categories. Similarly, we set the $\epsilon=0.3$ for image proxies collection and adopt the class balance strategy ~\cite{mmdetection}.

\begin{table}[t]
\resizebox{0.48\textwidth}{!}{
\begin{centering}
\begin{tabular}{c|c|ccc|cc}
\toprule

\multirow{2}*{Method} & \multirow{2}*{ZS} & \multicolumn{3}{c|}{15-way}  & \multicolumn{2}{c}{10-way}\tabularnewline
~ & ~ & 4-S & 8-S & 16-S & 10-S & 20-S\tabularnewline
\midrule
PointNet++ \cite{qi2017pointnet} & - & 41.0 & 47.6 & 55.0& - & -\tabularnewline
PointCLIP \cite{zhang2022pointclip} & 15.4 & 46.0 & 50.0 & 55.6& - & - \tabularnewline
Clip2Point \cite{huang2022clip2point} & 23.3 & - & -& - & - & -\tabularnewline
CrossPoint \cite{afham2022crosspoint} & - & - & - & - &  58.7$\pm$1.8 & 64.6$\pm$1.2\tabularnewline
\midrule
CLIP${^2}$ & \textcolor{blue}{39.1}  & \textcolor{blue}{51.3}  & \textcolor{blue}{59.6} & \textcolor{blue}{62.5}& \textcolor{blue}{60.6$\pm$2.5} & \textcolor{blue}{66.3$\pm$3.2} \tabularnewline
\bottomrule
\end{tabular}
\par\end{centering}
}
\vspace{-2mm}
\caption{\textbf{Zero-shot and Few-shot classification results on ScanObjectNN.} \textbf{ZS: }zero-shot. \textbf{K-way N-shot: }few-shot settings.}\label{tab-scanobj}
\end{table}

\paragraph{Quantity results. }
Since the outdoor point cloud is collected by LiDAR sensors, it has a wider perception range than RGB-D but leads to sparse distribution. Thus the projected depth representation of baselines results in severer information lost, as illustrated in the second row in Figure~\ref{fig-depth}. As shown in Table~\ref{tab-main-out}, our CLIP$^2$ considerably outperforms the baseline recognition results by more than 20\%, and our triplet proxies respectively boost two baselines by 9.5\% and 4.8\%. Additionally, we evaluate the localization ability on the outdoor scenario nuScenes in Table \ref{tab-det}. Due to the lack of works that tackle outdoor open-vocabulary localization problems, we follow classic detection accuracy metrics Precision(P.) and Recall(R.) as evaluation metrics. Specifically, we calculate the center distance between groundtruth bounding boxes and our 3D proxies that are predicted to belong to the same category as the groundtruth, and set the distance threshold as $\lambda=2$m. For those matched pairs that are closer than $\lambda$, we count the proxies as TPs. Otherwise, for those unmatched proxies and groundtruth, we count them as FPs and FNs respectively, thus P.=$\frac{\text{TPs}}{\text{TPs}+\text{FPs}}$, R.=$\frac{\text{TPs}}{\text{TPs}+\text{FNs}}$. As shown in Table~\ref{tab-det}, our CLIP$^2$ pipeline can provide high recall for outdoor objects. Since CLIP$^2$ is highly sensitive to open-world objects and can perceive categories beyond groundtruth list, it tends to create overmuch predictions thus the precision is comparably low. The perception ability of open-world objects can be viewed in Figure~\ref{fig-vis}\textcolor{red}{(b)}.

\paragraph{Quality results. }
We show off two outdoor scenes of nuScenes~\cite{caesar2020nuscenes} in  Figure~\ref{fig-vis}\textcolor{red}{(b-\romannumeral1)} and Figure~\ref{fig-vis}\textcolor{red}{(b-\romannumeral2)}. In addition to perceiving those common categories, our CLIP$^2$ surprisingly localizes and recognizes those uncommon 3D objects in 3D scenes such as the tires of vehicles, the plastic bag in the hand of pedestrian as well as the plastic bag on the road. We believe it contributes to auto-driving safety by providing the localization and recognition of universal obstacles to facilitate follow-up driving decisions.

\subsection{Few-shot Classification}
\paragraph{Setting. }
Lightweight few-shot learning is practical for application by finetuning the pretraining model with given limited data annotations, which can also validate the generalization capability of our learned representation. To make a fair comparison, we follow the existing methods~\cite{zhang2022pointclip, afham2022crosspoint} to conduct experiments under \textquotedblleft K-way N-shot\textquotedblright{} setting on the challenging realistic object-level dataset ScanObjectNN~\cite{uy2019revisiting}, where we randomly sample N point cloud objects from each of the randomly selected K classes.

\begin{table}[t]
	\begin{center}
	\resizebox{0.49\textwidth}{!}{
		\begin{centering}
		\begin{tabular}{ccc|ccc} 
\toprule
~&Rep. & Obj. & Avg.\_IN& Avg.\_OUT & Avg.\_OBJ\tabularnewline
\midrule 

(a)& Depth & Lang.-Depth  & 38.0  & 9.3 & 31.8   \tabularnewline
(b)& Depth & Image-Depth & 56.4 & 21.7 &  39.0 \tabularnewline
\midrule
(c)& PC. & Lang.-Point & 52.6 & 26.8  & 37.7  \tabularnewline
(d)& PC. & Image-Point  & 56.5  & 24.4 & 37.8 \tabularnewline
\midrule
(e)& PC. & Lang.-Image-Point  & \textcolor{blue}{61.3} & \textcolor{blue}{28.8} & \textcolor{blue}{39.4} \tabularnewline
\bottomrule
		\end{tabular}
	\end{centering}
		}
		\end{center}
					\vspace{-5mm}
	\caption{Ablations on representation learning. \textbf{Rep.: }the recognition representations. \textbf{Obj.: }the learning objectives. \textbf{PC.: }point cloud. \textbf{Lang.: }language space.
	}\label{tab-ablation-learning}
\vspace{-3mm}
\end{table}

\paragraph{Quantity results. }
As illustrated in Table~\ref{tab-scanobj}, we compare with representative 3D networks PointNet++\cite{qi2017pointnet++}, the recent zero-shot approach PointClip~\cite{zhang2022pointclip} as well as a state-of-the-art representation learning method CrossPoint~\cite{afham2022crosspoint}, which conducts contrastive pretraining between point cloud and rendered images on CAD dataset ShapeNet~\cite{chang2015shapenet}. As we can see, with a slight number of samples, our CLIP$^2$ can boost the classification results by a large margin,  exceeding PointClip by 5.3\%, 9.6\% and 6.9\% with 4, 8 and 16 shots. Besides, we outperform CrossPoint with considerable gain, illustrating our pretraining strategy on collected proxies can learn sufficient knowledge from realistic open world to generate transferable 3D representation, which is superior to the pretraining on a small-scale synthetic dataset.

\subsection{Ablations and Analysis }

\vspace{1mm}
\paragraph{Ablations on representation learning. }
To observe the transferability of different representations and the effect of different learning objectives, we conduct ablations and report the  mean average Top1 accuracy across all classes of zero-shot recognition in indoor scenarios~\cite{song2015sun} (Avg.\_IN), outdoor scenarios~\cite{caesar2020nuscenes} (Avg.\_OUT) and object-level benchmark~\cite{uy2019revisiting} (Avg.\_OBJ), which is shown in Table~\ref{tab-ablation-learning}. 
Firstly, for fair comparisons, we follow~\cite{zhang2022pointclip, huang2022clip2point} to project input point cloud into depth maps in $N_V$  different views as alternative representation. Secondly, we adopt various objectives to learn different correlation alignments across language, image and point cloud feature space or depth space. Specifically, comparing (a) and (b), aligning depth space to image space yields better transfer performance due to the similar data structure of image and depth map. In (c) and (d), point cloud representation is better when aligning to image space in indoor scenes, while better to align with language space in outdoor scenes due to the data discrepancy between image-like RGB-D points and sparse LiDAR points. Generally, 3D point cloud representation outperforms depth representation in all benchmarks due to preserving the complete 3D structure and sufficient 3D-specific information. Comparing (e) and (d), the joint alignment between three feature spaces contributes to the best 3D point cloud representation transferability on all benchmarks. 

\vspace{-2mm}
\paragraph{Analysis of representation ensembling. }
Intuitively, different representations contain different perspectives of knowledge, which can be potentially merged to achieve the optimum results during inference. To validate the ensembling application, we adopt three optional representation modals, \textit{i.e.} point clouds, projected depth maps and corresponding image patches, where depth representation is trained on our proxies and image representation is generated from pretrained image branch of CLIP. We ensemble their predicted logits by simple summation as the final output,  and illustrate the separate recognition results and ensembling performance in Table~\ref{tab-ablation-ensemble}. Benefiting from the sufficient knowledge learned from massive CLIP training data, image representation presents best performance in separate applications. By merging the complementary knowledge, our 3D representation leads to gains of 4.6\% indoors~\cite{song2015sun} and 2.8\% outdoors. Though further improving the indoor recognition performance with 0.9\% when merged, depth representation yeilds 1.6\% drop for outdoor objects, illustrating the information lost especially for outdoor scenarios. Since image representation is sometimes missing, such as in ~\cite{uy2019revisiting}, our 3D representation is more robust for 3D applications.


\begin{table}[t]
	\begin{center}
	\resizebox{0.48\textwidth}{!}{
		\begin{centering}
		\begin{tabular}{cccc|ccc} 
\toprule
~&PC. & Image & Depth  & Avg.\_IN& Avg.\_OUT & Avg.\_OBJ\tabularnewline
\midrule

(\romannumeral1)&$\checkmark$ &  &  & 61.3  & 28.8 & 39.4 \tabularnewline
(\romannumeral2)& & $\checkmark$ &  & 64.2  & 41.1 & -  \tabularnewline

(\romannumeral3)&&  & $\checkmark$  & 56.9  & 23.9 & 39.0 \tabularnewline
\midrule
(f)&$\checkmark$ & $\checkmark$ &  & 68.7  & \textcolor{blue}{43.9} &  - \tabularnewline
(g)&$\checkmark$ &  & $\checkmark$ & 64.8  & 30.4 & \textcolor{blue}{43.2}  \tabularnewline
(h)&$\checkmark$ & $\checkmark$ &  $\checkmark$  & \textcolor{blue}{69.6}  & 42.3 & - \tabularnewline
\bottomrule
		\end{tabular}
	\end{centering}
		}
		\end{center}
					\vspace{-5mm}
	\caption{Analysis on the representation ensembling schemes. 
	}\label{tab-ablation-ensemble}
 \vspace{-3mm}
\end{table}
\section{Limitation}
 As a pilot work for the language-3D pretraining problem, though CLIP$^2$ enables zero-shot localization and recognition with proposed triplet proxy generation and learned transferable 3D representation, it can not provide the accurate tight bounding box for open-world 3D objects as a common detector does. We believe CLIP$^2$ can facilitate the development of open-world 3D detectors by introducing the recognition ability to general 3D detectors or providing presented 3D proxies to enable further training of 3D detectors.
\section{Conclusion}
In this paper, we present a novel contrastive language-image-point cloud pretraining framework, CLIP$^2$, which consists of a triplet proxy collection scheme and a cross-modal contrastive learning mechanism. Based on the observation that realistic scenarios contain a massive amount of open-world objects, we innovatively propose to collect triplet proxies from realistic scenes as pretraining data. We then conduct cross-modal contrastive alignment across language, image and point cloud feature space to learn transferable 3D representation. The zero-shot transfer results on various indoor and outdoor benchmarks validate the ability of  CLIP$^2$ for 3D open-world understanding.

\paragraph{Acknowledgements} We gratefully acknowledge the support of MindSpore\footnote{\url{https://www.mindspore.cn/}}, CANN (Compute
Architecture for Neural Networks) and Ascend AI Processor in this work.

{\small
\bibliographystyle{./ieee_fullname}
\bibliography{reference}

\begin{thebibliography}{10}\itemsep=-1pt

\bibitem{afham2022crosspoint}
Mohamed Afham, Isuru Dissanayake, Dinithi Dissanayake, Amaya Dharmasiri,
  Kanchana Thilakarathna, and Ranga Rodrigo.
\newblock Crosspoint: Self-supervised cross-modal contrastive learning for 3d
  point cloud understanding.
\newblock In {\em CVPR}, 2022.

\bibitem{caesar2020nuscenes}
Holger Caesar, Varun Bankiti, Alex~H Lang, Sourabh Vora, Venice~Erin Liong,
  Qiang Xu, Anush Krishnan, Yu Pan, Giancarlo Baldan, and Oscar Beijbom.
\newblock nuscenes: A multimodal dataset for autonomous driving.
\newblock In {\em CVPR}, 2020.

\bibitem{chang2015shapenet}
Angel~X Chang, Thomas Funkhouser, Leonidas Guibas, Pat Hanrahan, Qixing Huang,
  Zimo Li, Silvio Savarese, Manolis Savva, Shuran Song, Hao Su, et~al.
\newblock Shapenet: An information-rich 3d model repository.
\newblock {\em arXiv preprint arXiv:1512.03012}, 2015.

\bibitem{mmdetection}
Kai Chen, Jiaqi Wang, Jiangmiao Pang, Yuhang Cao, Yu Xiong, Xiaoxiao Li,
  Shuyang Sun, Wansen Feng, Ziwei Liu, Jiarui Xu, Zheng Zhang, Dazhi Cheng,
  Chenchen Zhu, Tianheng Cheng, Qijie Zhao, Buyu Li, Xin Lu, Rui Zhu, Yue Wu,
  Jifeng Dai, Jingdong Wang, Jianping Shi, Wanli Ouyang, Chen~Change Loy, and
  Dahua Lin.
\newblock {MMDetection}: Open mmlab detection toolbox and benchmark.
\newblock {\em arXiv preprint arXiv:1906.07155}, 2019.

\bibitem{cheraghian2019mitigating}
Ali Cheraghian, Shafin Rahman, Dylan Campbell, and Lars Petersson.
\newblock Mitigating the hubness problem for zero-shot learning of 3d objects.
\newblock In {\em BMVC}, 2019.

\bibitem{cheraghian2020transductive}
Ali Cheraghian, Shafin Rahman, Dylan Campbell, and Lars Petersson.
\newblock Transductive zero-shot learning for 3d point cloud classification.
\newblock In {\em WACV}, 2020.

\bibitem{cheraghian2022zero}
Ali Cheraghian, Shafin Rahman, Townim~F Chowdhury, Dylan Campbell, and Lars
  Petersson.
\newblock Zero-shot learning on 3d point cloud objects and beyond.
\newblock {\em IJCV}, 2022.

\bibitem{cheraghian2019zero}
Ali Cheraghian, Shafin Rahman, and Lars Petersson.
\newblock Zero-shot learning of 3d point cloud objects.
\newblock In {\em MVA}. IEEE, 2019.

\bibitem{dai2017scannet}
Angela Dai, Angel~X Chang, Manolis Savva, Maciej Halber, Thomas Funkhouser, and
  Matthias Nie{\ss}ner.
\newblock Scannet: Richly-annotated 3d reconstructions of indoor scenes.
\newblock In {\em CVPR}, 2017.

\bibitem{ding2019votenet}
Zhipeng Ding, Xu Han, and Marc Niethammer.
\newblock Votenet: A deep learning label fusion method for multi-atlas
  segmentation.
\newblock In {\em MICCAI}, 2019.

\bibitem{gupta2019lvis}
Agrim Gupta, Piotr Dollar, and Ross Girshick.
\newblock Lvis: A dataset for large vocabulary instance segmentation.
\newblock In {\em CVPR}, pages 5356--5364, 2019.

\bibitem{ha2022semantic}
Huy Ha and Shuran Song.
\newblock Semantic abstraction: Open-world 3d scene understanding from 2d
  vision-language models.
\newblock In {\em CoRL}, 2022.

\bibitem{huang2022clip2point}
Tianyu Huang, Bowen Dong, Yunhan Yang, Xiaoshui Huang, Rynson~WH Lau, Wanli
  Ouyang, and Wangmeng Zuo.
\newblock Clip2point: Transfer clip to point cloud classification with
  image-depth pre-training.
\newblock {\em arXiv preprint arXiv:2210.01055}, 2022.

\bibitem{jia2021scaling}
Chao Jia, Yinfei Yang, Ye Xia, Yi-Ting Chen, Zarana Parekh, Hieu Pham, Quoc Le,
  Yun-Hsuan Sung, Zhen Li, and Tom Duerig.
\newblock Scaling up visual and vision-language representation learning with
  noisy text supervision.
\newblock In {\em ICML}. PMLR, 2021.

\bibitem{kingma2014adam}
Diederik~P Kingma and Jimmy Ba.
\newblock Adam: A method for stochastic optimization.
\newblock {\em arXiv preprint arXiv:1412.6980}, 2014.

\bibitem{li2021align}
Junnan Li, Ramprasaath Selvaraju, Akhilesh Gotmare, Shafiq Joty, Caiming Xiong,
  and Steven Chu~Hong Hoi.
\newblock Align before fuse: Vision and language representation learning with
  momentum distillation.
\newblock {\em NeurIPS}, 2021.

\bibitem{liang2021exploring}
Hanxue Liang, Chenhan Jiang, Dapeng Feng, Xin Chen, Hang Xu, Xiaodan Liang, Wei
  Zhang, Zhenguo Li, and Luc Van~Gool.
\newblock Exploring geometry-aware contrast and clustering harmonization for
  self-supervised 3d object detection.
\newblock In {\em ICCV}, 2021.

\bibitem{liu2022masked}
Haotian Liu, Mu Cai, and Yong~Jae Lee.
\newblock Masked discrimination for self-supervised learning on point clouds.
\newblock In {\em ECCV}, 2022.

\bibitem{lu2022open}
Yuheng Lu, Chenfeng Xu, Xiaobao Wei, Xiaodong Xie, Masayoshi Tomizuka, Kurt
  Keutzer, and Shanghang Zhang.
\newblock Open-vocabulary 3d detection via image-level class and debiased
  cross-modal contrastive learning.
\newblock {\em arXiv preprint arXiv:2207.01987}, 2022.

\bibitem{mao2021one}
Jiageng Mao, Minzhe Niu, Chenhan Jiang, Hanxue Liang, Jingheng Chen, Xiaodan
  Liang, Yamin Li, Chaoqiang Ye, Wei Zhang, Zhenguo Li, et~al.
\newblock One million scenes for autonomous driving: Once dataset.
\newblock {\em arXiv preprint arXiv:2106.11037}, 2021.

\bibitem{mao20223d}
Jiageng Mao, Shaoshuai Shi, Xiaogang Wang, and Hongsheng Li.
\newblock 3d object detection for autonomous driving: A review and new
  outlooks.
\newblock {\em arXiv preprint arXiv:2206.09474}, 2022.

\bibitem{misra2021end}
Ishan Misra, Rohit Girdhar, and Armand Joulin.
\newblock An end-to-end transformer model for 3d object detection.
\newblock In {\em ICCV}, 2021.

\bibitem{paigwar2021frustum}
Anshul Paigwar, David Sierra-Gonzalez, {\"O}zg{\"u}r Erkent, and Christian
  Laugier.
\newblock Frustum-pointpillars: A multi-stage approach for 3d object detection
  using rgb camera and lidar.
\newblock In {\em ICCV}, 2021.

\bibitem{pang2022masked}
Yatian Pang, Wenxiao Wang, Francis~EH Tay, Wei Liu, Yonghong Tian, and Li Yuan.
\newblock Masked autoencoders for point cloud self-supervised learning.
\newblock In {\em ECCV}, 2022.

\bibitem{paszke2017automatic}
Adam Paszke, Sam Gross, Soumith Chintala, Gregory Chanan, Edward Yang, Zachary
  DeVito, Zeming Lin, Alban Desmaison, Luca Antiga, and Adam Lerer.
\newblock Automatic differentiation in pytorch.
\newblock 2017.

\bibitem{phan2018dgcnn}
Anh~Viet Phan, Minh Le~Nguyen, Yen Lam~Hoang Nguyen, and Lam~Thu Bui.
\newblock Dgcnn: A convolutional neural network over large-scale labeled
  graphs.
\newblock {\em Neural Networks}, 2018.

\bibitem{qi2018frustum}
Charles~R Qi, Wei Liu, Chenxia Wu, Hao Su, and Leonidas~J Guibas.
\newblock Frustum pointnets for 3d object detection from rgb-d data.
\newblock In {\em CVPR}, 2018.

\bibitem{qi2017pointnet}
Charles~R Qi, Hao Su, Kaichun Mo, and Leonidas~J Guibas.
\newblock Pointnet: Deep learning on point sets for 3d classification and
  segmentation.
\newblock In {\em CVPR}, 2017.

\bibitem{qi2017pointnet++}
Charles~Ruizhongtai Qi, Li Yi, Hao Su, and Leonidas~J Guibas.
\newblock Pointnet++: Deep hierarchical feature learning on point sets in a
  metric space.
\newblock {\em NeurIPS}, 2017.

\bibitem{radford2021learning}
Alec Radford, Jong~Wook Kim, Chris Hallacy, Aditya Ramesh, Gabriel Goh,
  Sandhini Agarwal, Girish Sastry, Amanda Askell, Pamela Mishkin, Jack Clark,
  et~al.
\newblock Learning transferable visual models from natural language
  supervision.
\newblock In {\em ICML}, 2021.

\bibitem{rother2004grabcut}
Carsten Rother, Vladimir Kolmogorov, and Andrew Blake.
\newblock " grabcut" interactive foreground extraction using iterated graph
  cuts.
\newblock {\em TOG}, 2004.

\bibitem{schubert2017dbscan}
Erich Schubert, J{\"o}rg Sander, Martin Ester, Hans~Peter Kriegel, and Xiaowei
  Xu.
\newblock Dbscan revisited, revisited: why and how you should (still) use
  dbscan.
\newblock {\em TODS}, 42(3):1--21, 2017.

\bibitem{sharma2020self}
Charu Sharma and Manohar Kaul.
\newblock Self-supervised few-shot learning on point clouds.
\newblock {\em NeurIPS}, 2020.

\bibitem{shi2019pointrcnn}
Shaoshuai Shi, Xiaogang Wang, and Hongsheng Li.
\newblock Pointrcnn: {3D} object proposal generation and detection from point
  cloud.
\newblock In {\em CVPR}, 2019.

\bibitem{song2015sun}
Shuran Song, Samuel~P Lichtenberg, and Jianxiong Xiao.
\newblock Sun rgb-d: A rgb-d scene understanding benchmark suite.
\newblock In {\em CVPR}, 2015.

\bibitem{sun2020scalability}
Pei Sun, Henrik Kretzschmar, Xerxes Dotiwalla, Aurelien Chouard, Vijaysai
  Patnaik, Paul Tsui, James Guo, Yin Zhou, Yuning Chai, Benjamin Caine, et~al.
\newblock Scalability in perception for autonomous driving: Waymo open dataset.
\newblock In {\em CVPR}, 2020.

\bibitem{uy2019revisiting}
Mikaela~Angelina Uy, Quang-Hieu Pham, Binh-Son Hua, Thanh Nguyen, and Sai-Kit
  Yeung.
\newblock Revisiting point cloud classification: A new benchmark dataset and
  classification model on real-world data.
\newblock In {\em ICCV}, 2019.

\bibitem{vishwanath2009modelnet}
Kashi~Venkatesh Vishwanath, Diwaker Gupta, Amin Vahdat, and Ken Yocum.
\newblock Modelnet: Towards a datacenter emulation environment.
\newblock In {\em 2009 IEEE Ninth International Conference on Peer-to-Peer
  Computing}. IEEE, 2009.

\bibitem{wang2021unsupervised}
Hanchen Wang, Qi Liu, Xiangyu Yue, Joan Lasenby, and Matt~J Kusner.
\newblock Unsupervised point cloud pre-training via occlusion completion.
\newblock In {\em ICCV}, 2021.

\bibitem{xie2020pointcontrast}
Saining Xie, Jiatao Gu, Demi Guo, Charles~R Qi, Leonidas Guibas, and Or Litany.
\newblock Pointcontrast: Unsupervised pre-training for 3d point cloud
  understanding.
\newblock In {\em ECCV}, 2020.

\bibitem{yao2022detclip}
Lewei Yao, Jianhua Han, Youpeng Wen, Xiaodan Liang, Dan Xu, Wei Zhang, Zhenguo
  Li, Chunjing Xu, and Hang Xu.
\newblock Detclip: Dictionary-enriched visual-concept paralleled pre-training
  for open-world detection.
\newblock In {\em NeurIPS}, 2022.

\bibitem{ye2020sarpnet}
Yangyang Ye, Houjin Chen, Chi Zhang, Xiaoli Hao, and Zhaoxiang Zhang.
\newblock Sarpnet: Shape attention regional proposal network for lidar-based 3d
  object detection.
\newblock {\em Neurocomputing}, 2020.

\bibitem{yin2020center}
Tianwei Yin, Xingyi Zhou, and Philipp Kr{\"a}henb{\"u}hl.
\newblock Center-based 3d object detection and tracking.
\newblock {\em arXiv preprint arXiv:2006.11275}, 2020.

\bibitem{yu2022point}
Xumin Yu, Lulu Tang, Yongming Rao, Tiejun Huang, Jie Zhou, and Jiwen Lu.
\newblock Point-bert: Pre-training 3d point cloud transformers with masked
  point modeling.
\newblock In {\em CVPR}, 2022.

\bibitem{zhang2022point}
Renrui Zhang, Ziyu Guo, Peng Gao, Rongyao Fang, Bin Zhao, Dong Wang, Yu Qiao,
  and Hongsheng Li.
\newblock Point-m2ae: multi-scale masked autoencoders for hierarchical point
  cloud pre-training.
\newblock In {\em NeurIPS}, 2022.

\bibitem{zhang2022pointclip}
Renrui Zhang, Ziyu Guo, Wei Zhang, Kunchang Li, Xupeng Miao, Bin Cui, Yu Qiao,
  Peng Gao, and Hongsheng Li.
\newblock Pointclip: Point cloud understanding by clip.
\newblock In {\em CVPR}, 2022.

\bibitem{zhang2020h3dnet}
Zaiwei Zhang, Bo Sun, Haitao Yang, and Qixing Huang.
\newblock H3dnet: 3d object detection using hybrid geometric primitives.
\newblock In {\em ECCV}, 2020.

\end{thebibliography}
 }

\clearpage
\renewcommand{\thetable}{A\arabic{table}}
\renewcommand{\thefigure}{A\arabic{figure}}
\appendix

\begin{figure*}[!ht]
	\begin{center}
		\includegraphics[width=0.8\linewidth]{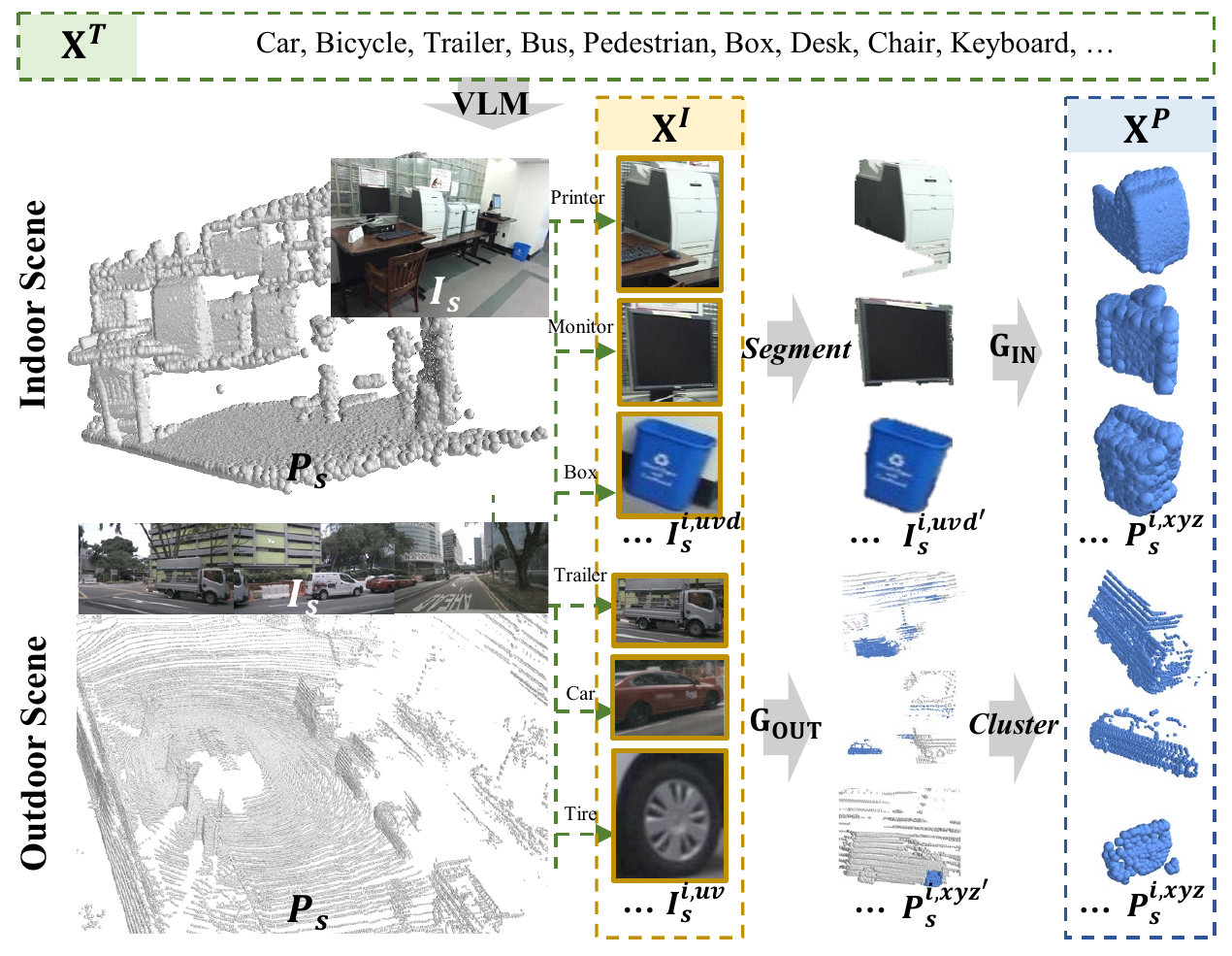}
	\end{center}
	\vspace{-5mm}
	\caption{
	Illustration of triplet proxy generation process.
	}
	\label{fig-proxy}
\vspace{2mm}
\end{figure*}

\section{Implement Details}
\subsection{Triplet Proxy Collection}
Generally, the triplet proxy collection concludes given text proxies $\textbf{\text{X}}^T$, extracting image proposals as 2D proxies $\textbf{\text{X}}^I$ and constructing 3D proxies $\textbf{\text{X}}^P$ with geometry relations between images and point clouds. According to the application scenarios, we detail the construction process in indoor and outdoor scenes separately. 
\paragraph{Indoor scenes. }
The indoor scenes $S$ usually adopt RGB-D sensors to collect images with corresponding depth maps as $I_s^{uvd}$, where $s\in |S|$. Specifically, we first provide given text proxy $\textbf{\text{X}}^T$ as the input of pretrained DetCLIP~\cite{yao2022detclip} to extract 2D image proposals $I_s^{i, uvd}, i\in |X_s^I|$, where $|X_s^I|$ denotes the amount of 2D proxies in scene $s$. Then we segment the foreground images with unsupervised GrabCut~\cite{rother2004grabcut} algorithm  as $I_s^{i, uvd^{'}}$ , thus the point cloud instances can be reconstructed by the RGB-D pixels with camera calibration $\text{G}_\text{IN}$, which can be formulated as:
\begin{equation*}
    \lceil x, y, z \rceil = \text{G}_\text{IN}^{-1} \times \lceil u, v, d \rceil,
\end{equation*}
where $\text{G}_\text{IN}=\text{I}\times\text{R}_c$ denotes the combination of the intrinsics matrix $\text{I}$ and the extrinsics matrix $\text{R}_c$ of RGB-D camera. 

\paragraph{Outdoor scenes. }
Considering a much wider perception range, outdoor scenes usually have LiDAR and camera sensors to capture point clouds $P_s^{xyz}$ and camera images $I_s^{uv}$. Thus point clouds can be projected into camera pixels with sensor transformation matrix $\text{G}_\text{OUT}$ as:
\begin{equation*}
    \lceil u, v, d \rceil = \text{G}_\text{OUT} \times \lceil x, y, z \rceil,
\end{equation*}
where $\text{G}_\text{OUT} = \text{I}\times \text{R}_{c}^{-1}\times \text{R}_{l}$ are the combination of camera intrinsics matrix $\text{I}$, camera extrinsics matrix $\text{R}_c$ and the LiDAR  extrinsics matrix $\text{R}_l$. Concretely, we first conduct a similar procedure to indoor scenes that produces 2D image proposals as $I_s^{i, uvd}, i\in |X_s^I|$ for 2D proxies $X_s^I$. Then we extract the 3D frustum $P_s^{i, xyz^{'}}$ by extruding the
2D image proposal into 3D space and conduct DBSCAN clustering within the frustum. Eventually, we obtain the 3D proxy instance by filtering the point cloud cluster $P_s^{i, xyz}$. The whole process of triplet proxy collection is illustrated in Figure~\ref{fig-proxy}.

\begin{table*}
\parbox{.47\linewidth}{
 \centering
     	\resizebox{.47\textwidth}{!}{
         \begin{tabular}{c|ccc}
			\hline
			Encoder & ScanNet & SUN RGB-D & ScanObjectNN\\
			\hline
			PointNet & 22.6 & 45.3 & 27.0 \\
			DGCNN & 26.0 & 52.7 & 34.0 \\
		    PointNet++  & \textcolor{blue}{38.5}  & \textcolor{blue}{61.3} & \textcolor{blue}{39.4} \\
			\hline
		\end{tabular}
		}
\caption{Comparison of point cloud encoders.}\label{tab-backbone}
}
\hfill
\parbox{0.53\linewidth}{
  \centering
	\resizebox{.53\textwidth}{!}{
		\begin{centering}
\begin{tabular}{c|cccc}
			\hline
            \multirow{2}*{Range} & \multicolumn{2}{c}{ScanNet}  & \multicolumn{1}{c}{SUN RGB-D} & \multicolumn{1}{c}{ScanObjectNN}\\
~& Main  Top1 & 384 cls. Top5& Main  Top1  & Top1\\
			\hline
			SUN=37 & 36.6&17.1& \textcolor{blue}{63.6}&34.4  \\
			LVIS=1203 & \textcolor{blue}{38.5} & \textcolor{blue}{22.0} & 61.3 & \textcolor{blue}{39.4} \\
			SCAN=384 &  39.5&23.0& 61.6&44.6\\
			\hline
		\end{tabular}
	       \end{centering}
		      }
\caption{Comparison with different proxy range.}\label{tab-proxy}
}
\end{table*}

\subsection{Contrastive Pretraining}
Our main paper applies the popular point cloud classifier PointNet++~\cite{qi2017pointnet++} as our point cloud encoder. Concretely, we use two set abstraction layers that aggregate multi-scale information and then encode the feature vectors for point cloud instances by three fully convolutional layers. We remove the convolutional head of PointNet++  since the point cloud features of CLIP$^2$ can be directly referenced to the language embedding for downstream tasks.

We conduct all experiments using Pytorch \cite{paszke2017automatic}, 8 Tesla V100 cards on a single server. We randomly sample 2048 points on each object both for training and testing. At training time, AdamW optimizer \cite{kingma2014adam} is performed on 8 GPUs with 200 batch sizes on each. The learning rate is set to 0.006, 3e-2 as weight decay, and 0.9 as momentum. And we adopt the cosine decay with 1000 iteration warm-up. For both indoor and outdoor datasets, we train 100 epochs.

\subsection{ScanNet Dataset}
Considering the ambiguous synonyms in the raw classes, like ``handrail", ``stair rail" and ``banister", we involve a data preprocessing step aimed at merging raw classes using WordNet\footnote{\url{https://wordnet.princeton.edu/}} synonyms.
Specifically, the official file \textit{scannetv2-labels.combined.tsv} provided by ScanNet is utilized to identify synonyms for the various classes. This process resulted in the merging of 290 classes, which included 86 classes that lacked synonyms.
To further refine the merged classes, the 86 classes were subjected to an additional merging step. This step involved merging them into existing synonyms based on the path similarity in WordNet. The decision to merge was guided by a predefined threshold, such that only classes with a path similarity score above the threshold were merged.

The outcome of the above-described process was a final set of 249 classes deemed suitable for open-vocabulary evaluation. These classes represented a more refined and comprehensive set of merged classes, facilitating more reasonable and consistent evaluations.

\section{Additional Results}
\subsection{Different Point Cloud Encoder}
We compare three alternatives of point cloud encoder, including PointNet~\cite{qi2017pointnet}, DGCNN~\cite{phan2018dgcnn} and PointNet++~\cite{qi2017pointnet++}, and report the class average Top1 accuracy of zero-shot recognition in Table~\ref{tab-backbone}. Specifically, PointNet encodes point cloud features with point-wise MLP and max-pooling, DGCNN applies EdgeConv to extract edge features and then ensemble the point cloud features, while PointNet++ adopts additional hierarchical feature learning based on PointNet to leverage neighborhoods at multiple scales. As illustrated in Table~\ref{tab-backbone}, PointNet++ outperforms the other two encoders on all benchmarks, showing its superiority in extracting effective point cloud features. We believe more advanced point cloud encoder architectures can further enhance our learned 3D representation of CLIP$^2$.

\subsection{Impact Analysis of Proxy}
\paragraph{Proxy range. }
As the prior knowledge of open-world vocabularies, we adopt the caption list in 2D open-world dataset LVIS~\cite{gupta2019lvis} to set text proxies without human annotations. In Table~\ref{tab-proxy}, we transfer the text proxies to the groundtruth list of segmentation annotations of the SUN RGB-D~\cite{song2015sun}, which presents the congruous vocabulary range of dataset annotations with less noise but a narrow vision of open-vocabulary. Results in Table~\ref{tab-proxy} demonstrate that the groundtruth proxy range can improve the intra-dataset recognition performance on SUN RGB-D by 2.3\% average Top1 Acc. However, the inter-dataset performance drops 1.9\% on ScanNet and 5.0\% on ScanObjectNN, and yields a 4.9\% drop of average Top5 Acc on the extended vocabularies of ScanNet. The overall results validate that the open-world vocabulary of text proxies benefits transferable 3D representation learning.

\begin{figure}[hb]
\centering
\includegraphics[scale=0.8]{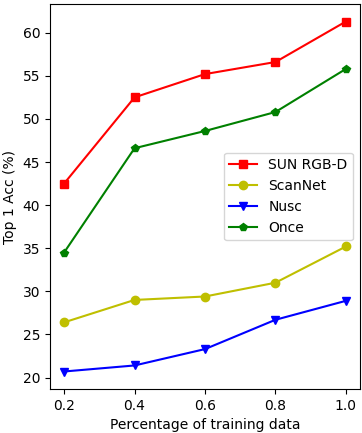}
\caption{Performance curves for training proxy quantity.}\label{fig-percentage}
\end{figure}

\paragraph{Proxy quantity. }
In Figure~\ref{fig-percentage}, we present the performance curves that demonstrate the consistency between the zero-shot recognition performance and increasing proxy data. Our analysis suggests that increasing the amount of training data in future work has the potential to further improve the upper bound of performance. These findings highlight the importance of scaling proxy data in a cost-effective manner.

\begin{table}
\resizebox{.46\textwidth}{!}{
\begin{centering}
\begin{tabular}{c|c|cc}
\toprule 
Method & Backbone & SUN & ScanNet\tabularnewline
\midrule
supervised L\_Head & \multirow{3}{*}{PointNet\cite{qi2017pointnet}} & 48.5 & -\tabularnewline
supervised T\_Head &  & 44.2 & 14.4\tabularnewline
\cmidrule{1-1} \cmidrule{3-4} \cmidrule{4-4} 
CLIP$^{2}$ &  & 45.3 & \textcolor{blue}{22.6}\tabularnewline
\midrule
\midrule 
supervised L\_Head &  & 63.4 & -\tabularnewline
supervised T\_Head & PointNet++ & 60.3 & 25.5\tabularnewline
\cmidrule{1-1} \cmidrule{3-4} \cmidrule{4-4} 
PointCLIP \cite{zhang2022pointclip} & \cite{qi2017pointnet++} & 11.5 & 6.7\tabularnewline
CLIP$^{2}$ &  & 61.3 & \textcolor{blue}{38.5} \tabularnewline
\bottomrule
\end{tabular}
\par\end{centering}
}
\caption{\label{tab:exam3}Comparisons with supervised baselines. We train supervised baselines on SUN RGB-D (SUN) dataset and evaluate recognition results on SUN and zero-shot performance on ScanNet. \textbf{L\_Head} and \textbf{T\_Head} indicate logit classification head and text classification head respectively.}
\end{table}

\subsection{Comparison with Supervised Baselines}
We conduct supervised training with popular 3D encoder PointNet~\cite{qi2017pointnet} and PointNet++ \cite{qi2017pointnet++} using annotations from SUN RGB-D training set. We consider two different settings as supervised baselines: 1) Traditional logit classification head \textbf{L\_Head}, which is fixed to the predefined training classes and fails to identify novel classes. 2) Text classification head indicated as \textbf{T\_Head}. Specifically, we replace the logit classification head with CLIP text embeddings. And the maximum cosine similarities between the 3D feature and text embeddings are the final results. According to the text classification head, we can compare the generalization of flexible categories with supervised training.

The results in Table~\ref{tab:exam3} show that CLIP$^2$ is comparable to supervised baselines on SUN RGB-D and outperforms them on ScanNet, illustrating the effectiveness of our unsupervised approach and its superiority in open-vocabulary understanding.


\section{More Qualitative Results}
\begin{figure*}[t]
\centering
\includegraphics[width=\linewidth]{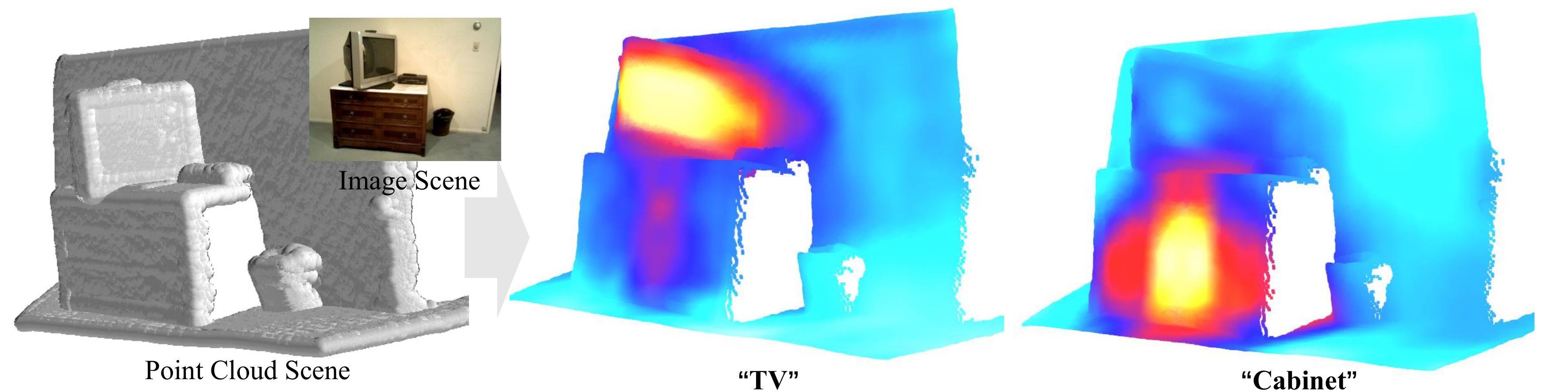}
\vspace{-2mm}
\caption{Saliency maps between texts and point cloud scenes. }
\label{fig-saliency}
\end{figure*}

\paragraph{Sailency map between text prompt and point cloud. }
To validate  our CLIP$^2$, we show a saliency map between the given text prompt and the point cloud within one scene in Figure~\ref{fig-saliency}. Specifically, we calculate the feature distances between the class texts and the point cloud scene and plot the saliency map, with lighter highlights representing smaller feature distances. The text feature has greater similarity to the point feature of the corresponding class, indicating the feature alignment between text and point cloud.

\begin{figure*}[t]
\vspace{4mm}
	\begin{center}
		\includegraphics[width=1\linewidth]{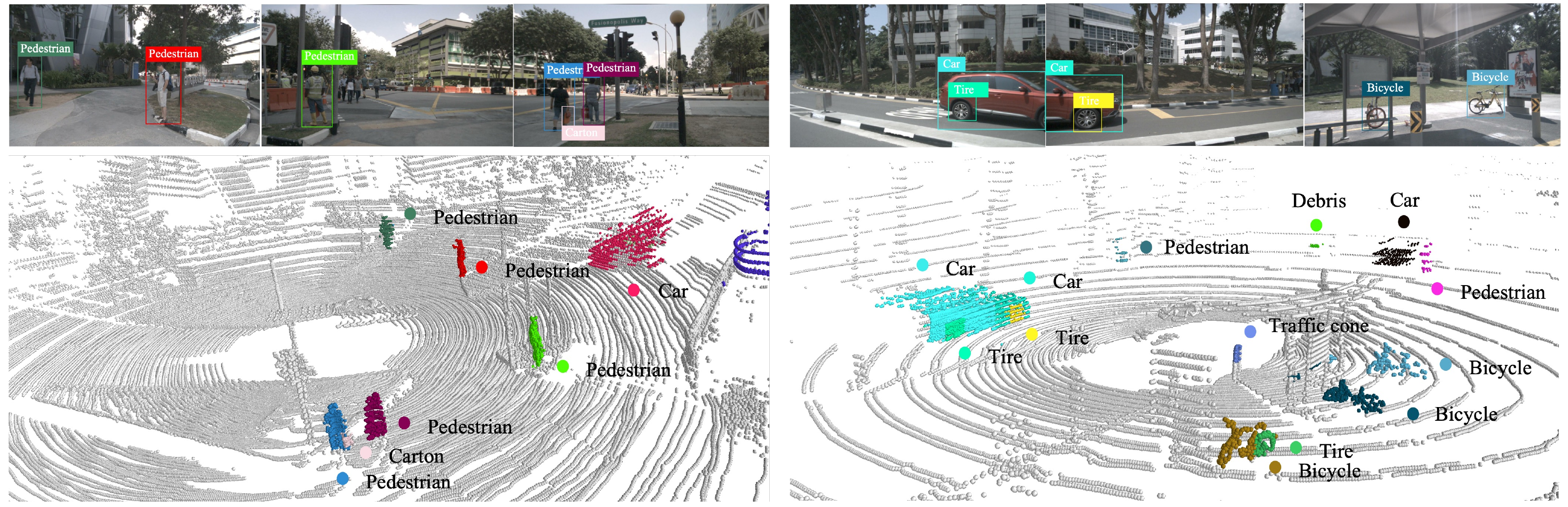}
		\includegraphics[width=1\linewidth]{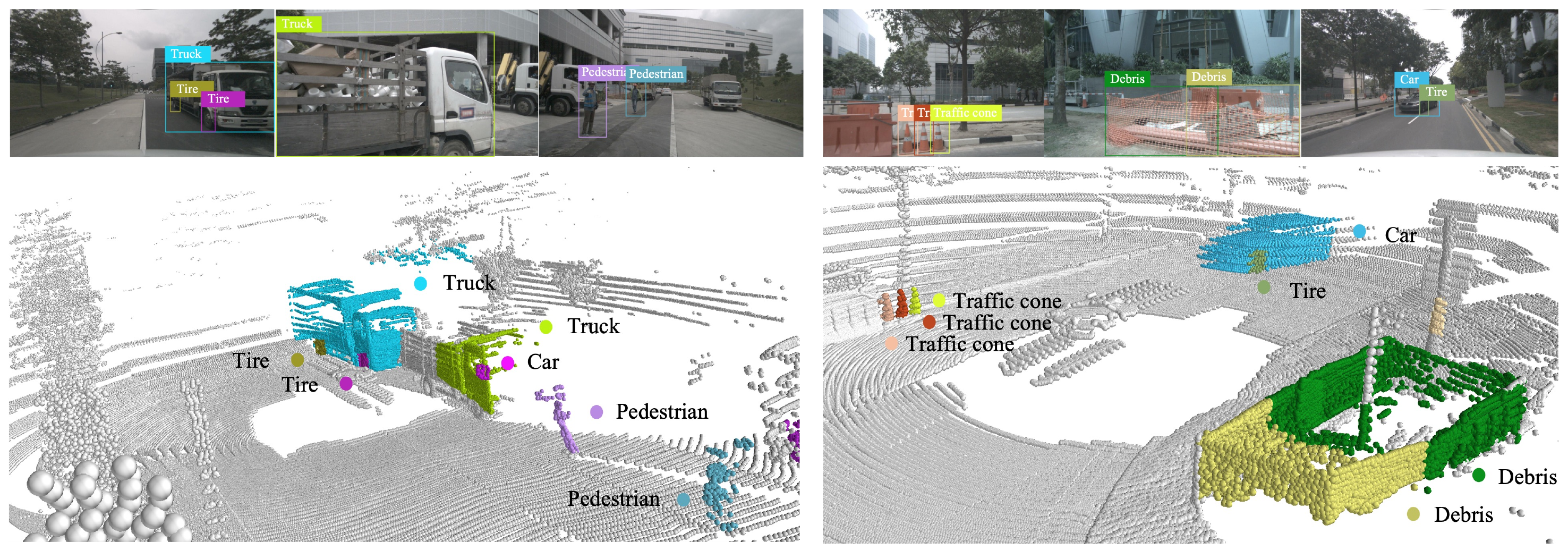}
	\end{center}
	\caption{
	More Visualizations of the zero-shot localization and recognition on the nuScenes dataset. The proposed CLIP$^2$ enables the open-world vocabularies beyond groundtruth annotations without extra human supervision, such as 'Tire' and ’Debris'. Best viewed in colors.
	}
	\label{fig-vis1}
\end{figure*}

\begin{figure*}[t]
	\begin{center}
		\includegraphics[width=1\linewidth]{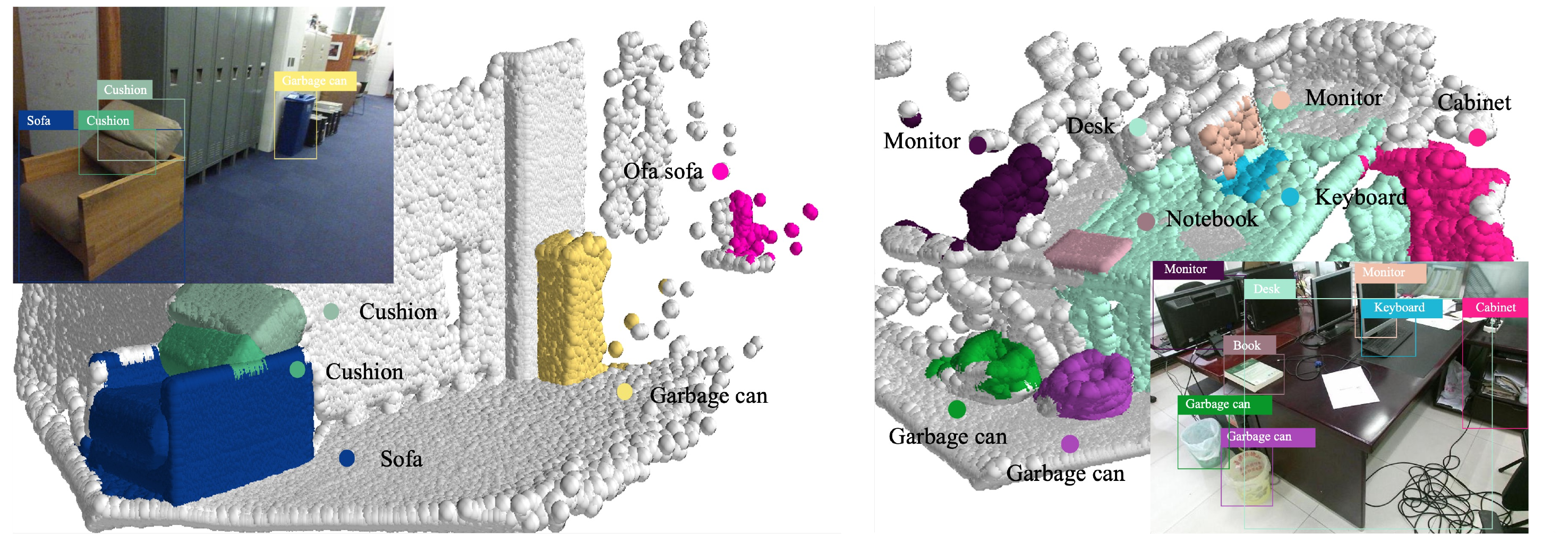}
		\includegraphics[width=1\linewidth]{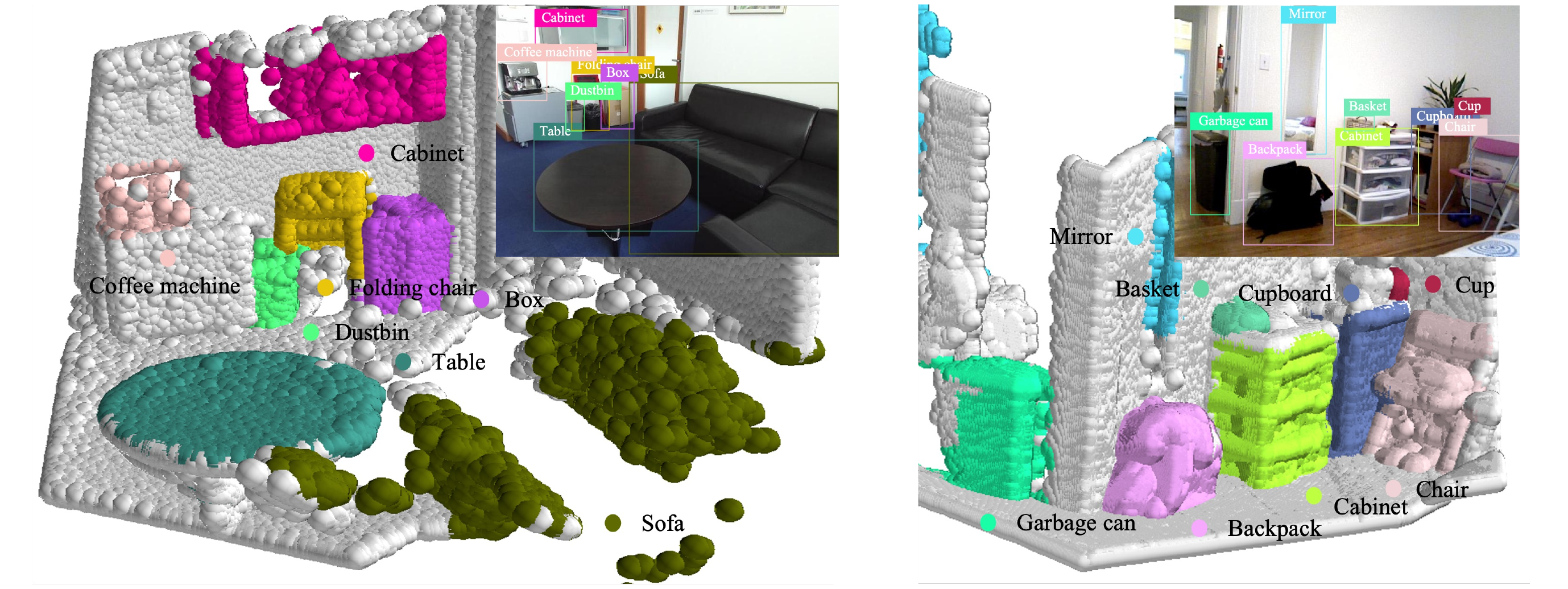}
	\end{center}
	\caption{
	More Visualizations of the zero-shot localization and recognition on SunRGB-D dataset. The proposed CLIP$^2$ shows open-world recognition ability in realistic scenarios. Best viewed in colors.
	}
	\label{fig-vis2}
\end{figure*}

\paragraph{Visualization of zero-shot localization. }
We show more qualitative results in Figure~\ref{fig-vis1} for indoor scenes SUN RGB-D \cite{song2015sun} and Figure~\ref{fig-vis2} for outdoor scenes nuScenes \cite{caesar2020nuscenes}. The visualization results illustrate the zero-shot localization and recognition abilities of CLIP$^2$. Specifically, the proposed CLIP$^2$ enables the open-world vocabularies beyond groundtruth annotations without extra human supervision, such as 'Tire' and ’Debris' in Figure~\ref{fig-vis1}.

\end{document}